\documentclass{article} 
\usepackage{iclr2027_conference,times}
\usepackage{graphicx} 

\usepackage{amsmath,amsfonts,bm}

\def\eqref#1{equation~\ref{#1}}

\def\1{\bm{1}}

\DeclareMathAlphabet{\mathsfit}{\encodingdefault}{\sfdefault}{m}{sl}
\SetMathAlphabet{\mathsfit}{bold}{\encodingdefault}{\sfdefault}{bx}{n}

\usepackage{hyperref}
\usepackage{url}
\usepackage{amsthm}
\usepackage{algorithm}
\usepackage{algpseudocode}
\usepackage{multirow}
\usepackage{multicol}
\usepackage{enumitem}
\usepackage{longtable}

\algrenewcommand\algorithmicrequire{\textbf{Input: }}
\algrenewcommand\algorithmicensure{\textbf{Output: }}
\usepackage{booktabs}

\theoremstyle{definition}
\newtheorem{definition}{Definition}

\title{Adaptive Multi-Value Control in LLMs via Causal Activation Steering}

\author{%
\begin{tabular}{c}
\\\\\\\\
\hspace*{0.6in}\bf Payel Bhattacharjee \qquad Ravi Tandon \\
\hspace*{0.9in}\normalfont School of Electrical, Computing, and Software Engineering\\
\hspace*{0.9in}\normalfont University of Arizona, Tucson, AZ, USA \\
\hspace*{1in}\normalfont Email: \texttt{\{payelb,tandonr\}@arizona.edu}
\end{tabular}}

\iclrfinalcopy 
\begin{document}

\maketitle

\begin{abstract}
Large language models (LLMs) are increasingly deployed in settings where responses must reflect multiple, potentially interacting social norms and human values. Activation steering offers a lightweight alternative to training-based alignment by modifying internal activations at inference time. However, prior human-value steering methods have largely considered values in isolation, while direct composition of multiple directions relies on fixed intervention strengths that cannot respond to the model's evolving internal state. Motivated by this key observation, we introduce \texttt{AIMES}, a framework for adaptive multi-value activation steering. \texttt{AIMES} constructs layer-specific bipolar directions for moral-foundation values and uses intermediate-layer vocabulary readouts as online observers. An observer-guided controller then adapts the strength of each requested value intervention at every decoding step based on its current observed state, without training a separate value-state estimator. Across multiple instruction-tuned model families, value combinations, and intervention depths, we find that multi-value controllability varies across both value combinations and intervention locations. Compared with fixed joint steering and prompt-based steering, \texttt{AIMES} shows depth-dependent advantages that are broadly supported across two independent evaluators, with some variation in the precise depth at which specific control effects emerge. These advantages come with smaller realized activation-space interventions than fixed-joint steering and comparable response quality. Overall, our results suggest that online observer feedback can provide lightweight, state-aware adaptation for single-pass multi-value steering.
\end{abstract}

\section{Introduction}

Large Language Models (LLMs) are increasingly deployed across critical domains such as education \citep{al2024analysis_education,alhafni2024llms_education}, research
\citep{ren2025towards_research,liao2024llms_RESEARCH}, healthcare \citep{yang2023large_healthcare,cascella2023evaluating_healthcare}, and finance \citep{lakkaraju2023llms_finance,zhao2024revolutionizing_finance}, where model behavior may need to reflect multiple, potentially interacting human values and social norms. This challenge is closely related to pluralistic alignment, which seeks to accommodate diverse and potentially competing values and perspectives, including through steerable adaptation to specified value trade-offs \citep{sorensen2024roadmap}. Existing alignment methods include reinforcement-learning-based approaches such as RLHF \citep{RLHFouyang2022training}, RLAIF \citep{RLAIFlee2023rlaif}, TRPO \citep{TRPOschulman2015trust}, and PPO \citep{PPOschulman2017proximal}, as well as reward-model-free methods such as DPO \citep{DPOrafailov2023direct}. However, these approaches generally require additional preference collection, optimization, or fine-tuning, making post-hoc adaptation to changing value requirements costly.

Activation steering \citep{subramani2205extracting,turner2023steering,zou2023representation,li2023inference} provides a lightweight inference-time alternative mechanism for controlling model behavior by modifying internal representations without updating model parameters. Recent work has extended this idea to human values, showing that moral concepts can be captured by layer-specific activation directions and causally influenced through intervention \citep{yu2026tracing}, but primarily considers one value at a time. Jointly steering multiple values is more challenging because their representations need not be independent: directions may be correlated or conflicting, and intervention along one value can affect the expression of others. Similar interference has been observed in generic multi-attribute activation steering, where composing several directions can reduce control reliability \citep{vanderweij2024extending,nguyen2025matsteer,ghasemi2026orbit,jiang2025msrs}. Existing approaches to multi-value alignment largely address this problem through other mechanisms, such as prompting \citep{kim2026valueflow} or combining separately generated value-conditioned outputs \citep{zheng2026vispa}. As a result, joint activation-level control of multiple human values remains comparatively underexplored.
\begin{figure}[t]
    \centering
    \includegraphics[scale=0.27]{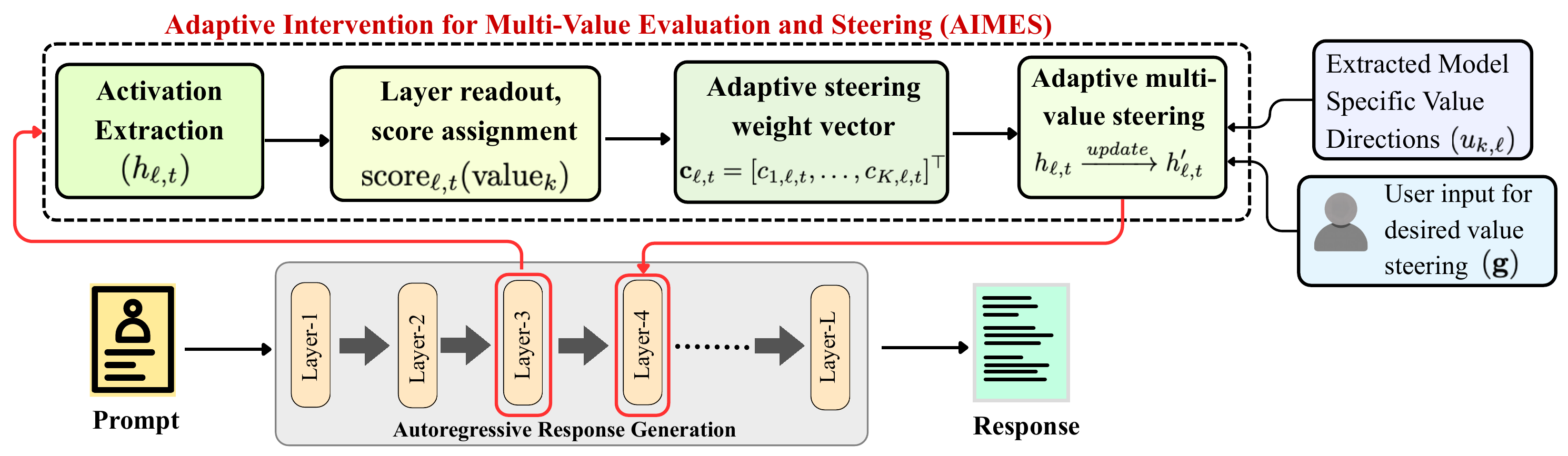}
    \caption{\textbf{Overview of \texttt{AIMES}.} At each decoding step, the hidden state activation $(h_{\ell,t})$ at the intervention layer is mapped to value-specific observer scores, which determine adaptive steering weights $(\mathbf{c}_{\ell,t})$ for the requested values. These weights modulate precomputed model and layer-specific value directions $(u_{k,\ell})$ before generation continues, enabling closed-loop multi-value steering.}
    \label{fig:aimes_workflow}
\end{figure}
A natural extension to multiple values is to jointly apply several value directions with fixed strengths, but such open-loop steering cannot adapt to
changes in the model's internal state during the generation process. Recent work instead formulates activation steering as feedback control, using estimates of the
current concept state to adjust intervention strength
\citep{nguyen2025feedback,bharadwaj2025stupid,prokopiou2026closing,skifstad2026local}. Nevertheless, these methods typically rely on separately trained probes, classifiers, or dynamical models. Intermediate-layer vocabulary readouts, including \textit{Logit Lens} \citep{nostalgebraist2020logitlens}, \textit{Tuned Lens} \citep{belrose2023tuned}, and \textit{J-Lens} \citep{gurnee2026jlens}, offer a lightweight alternative by exposing information directly from hidden representations. However, representational accessibility does not necessarily imply causal steerability \citep{billa2026predicting,nadaf2026steerable}, motivating an empirical comparison of candidate readouts before using them as feedback signals for adaptive control.

Motivated by these observations, we introduce \texttt{AIMES}
(\textbf{A}daptive \textbf{I}ntervention for \textbf{M}ulti-Value
\textbf{E}valuation and \textbf{S}teering), a framework that turns joint multi-value activation steering into a state-aware feedback process (Figure~\ref{fig:aimes_workflow}). Rather than composing multiple value directions with fixed strengths, \texttt{AIMES} observes their current expression in an intermediate representation and adjusts their relative strength during decoding. This enables adaptive coordination of interacting value directions without training a separate value-state estimator. Across multiple models, value objectives, and intervention depths, our results
show that multi-value controllability depends on both value combination and intervention location, and the cross-judge analysis reveals robust advantages in some depth regions. Our contributions are:
\begin{itemize}
    \item \textit{\textbf{Bipolar value directions and online observation.}}
    Building on contrastive activation steering \citep{rimsky-etal-2024-steering,yu2026tracing}, we construct model- and layer-specific bipolar human-value directions for bidirectional value control. Intermediate-layer vocabulary readouts are used as online value-state observers without training a separate state estimator.

    \item \textit{\textbf{Adaptive multi-value steering.}}  We formulate joint value steering as a closed-loop control problem and
    introduce an observer-guided controller that adapts the relative strength of multiple value-specific interventions at every decoding step.

    \item \textit{\textbf{Depth-aware and cross-judge empirical evaluation.}} Across five models spanning three families and parameter scales, and multiple multi-value objectives, we systematically evaluate adaptive control across intervention depth under two independent judges, \texttt{GPT-5.6-Sol} and \texttt{Claude-Opus-4.8}, characterizing the robustness and evaluator sensitivity of the observed multi-value control patterns.
\end{itemize}
\section{Related Work}
\label{sec:related}
In this section, we review prior work on activation steering and vocabulary
readout methods, and additional background is provided in
Appendix~\ref{app:prelim_background}.
\paragraph{Activation Steering for Human Values and Multiple Attributes.} Activation steering \citep{subramani2205extracting} modifies model behavior at inference time by intervening along directions identified in hidden representation space, without updating model parameters \citep{turner2023steering,zou2023representation,li2023inference}.
In the human-value setting, \citep{yu2026tracing} construct layer-specific contrastive directions for the five dimensions of Moral Foundations Theory \textit{ (MFT)}
\citep{graham2013MFT} and show that interventions along these directions can causally alter foundation-relevant behavior. Importantly, they also observe
cross-value effects, suggesting that value directions cannot generally be treated as independent. This issue becomes more pronounced when multiple directions are applied jointly. More broadly, multi-attribute steering methods such as \textit{MAT-Steer} \citep{nguyen2025matsteer}, \textit{ORBIT} \citep{ghasemi2026orbit}, \textit{MSRS} \citep{jiang2025msrs}, and \textit{K-Steering} \citep{oozeer2025ksteering} address interference among jointly controlled attributes using mechanisms including gating, orthogonalization, subspace decomposition, and nonlinear direction construction. These methods are motivated in part by evidence that naive vector composition becomes less reliable as the number of controlled attributes increases
\citep{vanderweij2024extending}. However, the intervention is typically fixed before generation and does not adapt to evolving attribute expression during decoding.
\paragraph{Steering Multiple Human Values.}
Recent work has explored multi-value control through prompting,
activation-level, and compositional approaches: \textit{VALUEFLOW}
\citep{kim2026valueflow} composes natural-language value specifications and
shows that value interactions can be reinforcing or competing, while
\textit{VISPA} \citep{zheng2026vispa} combines separately generated
value-conditioned outputs rather than jointly steering multiple directions
within a single generation. \citep{yu2026tracing} construct moral-foundation activation directions but
intervene on values individually. Joint activation-level control of multiple human values within a single generation
therefore remains comparatively underexplored. Inspired by these approaches, we compare \texttt{AIMES} with two non-adaptive alternatives:
\textit{Fixed Multi-Value Steering}, which jointly applies multiple value directions with predetermined strengths, and \textit{prompt steering}, which
specifies the same multi-value objective through natural-language instructions without modifying internal activations. Fixed joint steering is
closely related to generic multi-attribute activation steering, where vector composition can introduce interference and reduce controllability
\citep{vanderweij2024extending,nguyen2025matsteer,ghasemi2026orbit,
jiang2025msrs}. Concurrent work by \citep{pan2026spillover} attributes such
interference to geometric entanglement and proposes a static Gram-matrix correction. In contrast, \texttt{AIMES} uses online observer feedback to
adapt value-specific intervention strengths during generation.
\paragraph{Closed-Loop Activation Steering.}
Recent work formulates activation steering as a feedback-control problem, adapting intervention strength to the model's evolving internal state.
\citep{nguyen2025feedback} distinguish fixed-strength open-loop steering from
feedback-based control using the discrepancy between desired and observed concept states; \citep{bharadwaj2025stupid} use PID-style feedback with a chunk-level classifier during reasoning; \citep{wang2025adaptive} adapt truthfulness-steering strength online using probe-estimated activation states across different hallucination categories; \citep{prokopiou2026closing} study
feedback-controlled multi-attribute steering in symbolic music generation; and
\citep{skifstad2026local} model layer-wise computation as a locally linear dynamical system and derive interventions using linear-quadratic regulation. Relatedly, \textit{ODESteer} \citep{zhao2026odesteer} makes the steering
direction state-dependent through the gradient of a learned density-ratio barrier, but focuses on single-attribute control and requires fitting the
barrier. A key requirement in observer-based closed-loop steering is an \textit{observer} that estimates the controlled state, typically through a
separately trained classifier, probe, or fitted dynamical model \citep{nguyen2025feedback,bharadwaj2025stupid,
prokopiou2026closing,skifstad2026local}. \texttt{AIMES} instead uses intermediate vocabulary readouts as online observers and adapts the strengths of multiple jointly applied value directions during generation.

\paragraph{Intermediate-Layer Vocabulary Readouts.}
\label{related:readouts} To interpret the intermediate states, intermediate-layer readout methods map transformer hidden states into vocabulary-aligned coordinates: the \textit{Logit Lens} \citep{nostalgebraist2020logitlens}
directly applies the output projection, the \textit{Tuned Lens} \citep{belrose2023tuned} learns layer-specific transformations toward the final
output space, and \textit{J-Lens} \citep{gurnee2026jlens} uses layer-specific
Jacobian information to approximate downstream vocabulary-aligned content.
Beyond interpretability, \citep{billa2026predicting} show that Logit-Lens-based accessibility can help predict concept steerability and
effective intervention depth across models and concept families. However, representational visibility does not necessarily imply causal controllability: effective steering directions can remain weakly exposed by intermediate readouts such as the Logit Lens \citep{billa2026predicting,nadaf2026steerable}. This distinction is especially important in closed-loop steering, where the readout itself can potentially serve as the feedback signal for intervention.

\section{\texttt{AIMES}: Observer-Guided Adaptive Multi-Value Steering}
\label{sec:method}
In this section, we present our main proposed framework, \texttt{AIMES} (\textbf{A}daptive \textbf{I}ntervention for \textbf{M}ulti-Value \textbf{E}valuation and \textbf{S}teering), which consists of an offline value
direction-construction stage and an online adaptive-steering stage. Offline, we construct model and layer-specific human-value directions \(u_{k,\ell}\) for each target value \(V_k\) \citep{rimsky-etal-2024-steering,yu2026tracing}. During generation, intermediate-layer observers
\citep{gurnee2026jlens,nostalgebraist2020logitlens}
estimate the current expression of the controlled values and guide the allocation of value-specific steering strengths. Relative to Fixed Multi-Value
Steering, \texttt{AIMES} replaces fixed coefficients with observer-guided adaptive weights that are updated online. The \texttt{AIMES} framework has three major components: (1) \textit{value-specific steering directions}, which define the intervention axes; (2) an \textit{intermediate-layer value observer}, which estimates the current value state from a vocabulary-space readout; and (3) an \textit{observer-guided controller}, which adjusts the contribution of each requested value direction during generation. The intervention layer is fixed
within each generation run and varied across experiments to study the effect of intervention depth.
\subsection{Value-Specific Steering Direction Estimation}
\label{sec:value_directions}
\begin{figure}[!ht]
    \centering
    \includegraphics[scale=0.3]{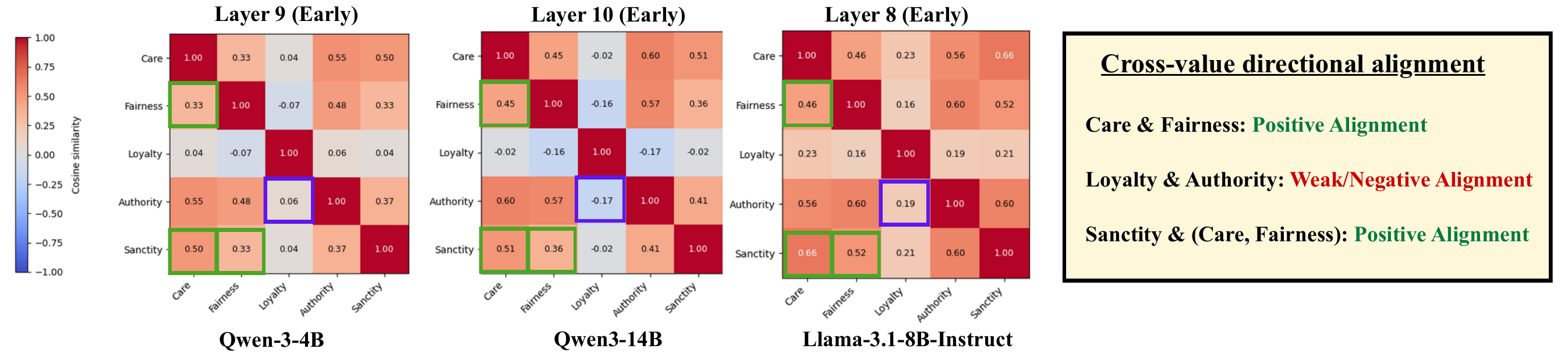}
    \caption{\textbf{Cross-value alignment analysis.} The figure shows early-layer cosine similarities for three representative models, highlighting positive Care-Fairness alignment, weaker Loyalty-Authority alignment, and positive Sanctity alignment with Care and Fairness.}
    \label{fig:early-value-direction}
\end{figure}
Given an autoregressive language model \(M\) with \(L\) layers, let
\(\ell\in\{1,\ldots,L\}\) denote the layer at which an intervention is applied during a given generation run. For an input prompt \(x_i\), the model generates a response
\(y_i=(y_i^1,\ldots,y_i^{T_i})\), where \(T_i\) is the number of generated tokens. We consider a set of \(K\) target human values \(\{V_1,\ldots,V_K\}\) and construct each value-specific steering direction from a matched contrastive corpus, following the mean-difference principle commonly used in activation steering \citep{rimsky-etal-2024-steering,yu2026tracing}. For each value \(V_k\), we form paired sets \(\mathcal D_k^{+}\) and \(\mathcal D_k^{-}\), where each
pair describes the same underlying scenario while expressing opposite poles of
the target value dimension. For the Moral Foundations Theory dimensions
considered here \citep{graham2013MFT}, these contrasts are \textit{Care/Harm,
Fairness/Cheating, Loyalty/Betrayal, Authority/Subversion, and
Sanctity/Degradation}. The complete vignette-generation, pairing, and validation
procedure is described in Appendix~\ref{app:contrastive_generation}. Formally, at layer $\ell$, we compute the positive and negative mean activations, $\mu_{k,\ell}^{+}$ and $\mu_{k,\ell}^{-}$, respectively, and define the normalized value direction $u_{k,\ell}$ as:
\begin{equation}
\small
\mu_{k,\ell}^{+} = \frac{1}{|\mathcal D_k^{+}|}
\sum_{x\in\mathcal D_k^{+}}
h_{\ell}(x),\qquad\mu_{k,\ell}^{-}=
\frac{1}{|\mathcal D_k^{-}|}
\sum_{x\in\mathcal D_k^{-}}
h_{\ell}(x), \qquad u_{k,\ell}
=
\frac{
\mu_{k,\ell}^{+}-\mu_{k,\ell}^{-}
}{
\left\|
\mu_{k,\ell}^{+}-\mu_{k,\ell}^{-}
\right\|_2
}.
\label{eq:value_direction}
\end{equation}

Since the positive and negative sets consist of one-to-one matched pairs of
equal size, the mean-difference direction is equivalent to the average
pairwise activation difference. The resulting direction $u_{k,\ell}$ therefore defines a bipolar activation-space axis, with its positive and negative orientations
corresponding to the two poles of value $V_k$. Following \textit{Contrastive
Activation Addition (CAA)} \citep{rimsky-etal-2024-steering}, this matched
construction provides a direct bidirectional intervention axis while reducing variation due to topic, actors, setting, and writing style. This is particularly useful for \texttt{AIMES}, which supports both amplification and suppression along the same value dimension. However, the existence of a separable direction does not imply that its
behavior is identical across layers. We therefore evaluate multi-value
steering across multiple intervention depths in Section~\ref{sec:experiments}.
\paragraph{Geometry-Guided Value-Objective Selection.}
For the multi-value steering analysis, we select three objectives to represent distinct interaction regimes suggested
by the learned value geometry. As illustrated in Figure \ref{fig:early-value-direction}, \textit{Care} and \textit{Fairness} are positively
aligned, motivating
$(\uparrow\mathrm{Care}\uparrow\mathrm{Fairness})$ as an aligned
coordination setting. \textit{Loyalty} and \textit{Authority} show substantially weaker and, in some models, negative alignment, motivating the
$(\uparrow\mathrm{Loyalty}\uparrow\mathrm{Authority})$ objective as a weakly-coupled
setting. These two pairings have also been considered in prior multi-value MFT steering \citep{kim2026valueflow} and are consistent with the broader
individualizing-binding distinction in MFT \citep{graham2011mapping,graham2013MFT}. Finally, \textit{Sanctity} is positively
aligned with both \textit{Care} and \textit{Fairness}; we therefore consider
$(\uparrow\mathrm{Care}\uparrow\mathrm{Fairness}\downarrow\mathrm{Sanctity})$ as a competing setting, where a value geometrically coupled to the promoted pair is intentionally steered in the opposite direction. Full similarity matrices across five models and depth regions are provided in Appendix~\ref{app:value_geometry}.

\subsection{Intermediate-Layer Value Observer: Design and Implementation}
\label{sec:observer} To adapt steering strength during generation, \texttt{AIMES} requires an estimate of how strongly each target value \(V_k\) is expressed in the model's current intermediate state. Rather than training a separate value classifier,
we derive this estimate directly from intermediate-layer vocabulary readouts.
We consider existing vocabulary readout methods \citep{gurnee2026jlens,nostalgebraist2020logitlens} as candidate observers, and use J-Lens as the default based on the ablation in Appendix~\ref{app:observer_ablation}. No
value-specific estimator is trained: the readout is fixed for a given model
and layer and is shared across all values and steering objectives. Logit Lens uses the model's unembedding matrix \(W_U\), while J-Lens additionally uses a
precomputed layer-specific Jacobian \(J_\ell\); further details are provided in Appendix~\ref{app:readouts}. For a given readout, let
\(z_{\ell,t}\in\mathbb{R}^{|\mathcal V|}\) denote its vocabulary-space output
at layer \(\ell\) and generation step \(t\). For each value \(V_k\), we define
a fixed token set \(\mathcal T_k\subset\mathcal V\) containing surface forms associated with that value (see Appendix \ref{app:aimes_algorithm} for details), following \textit{Moral Foundations Theory (MFT)} terminology \citep{graham2013MFT}. These sets are fixed before evaluation and are
independent of model activations. Extending them with additional
value-specific terms from external lexicons \citep{atari2023mft} may improve
observer sensitivity and is left for future work.

Vocabulary-space readouts are known to expose evolving intermediate predictions \citep{nostalgebraist2020logitlens,belrose2023tuned}, and prior work has shown that such projections can reveal interpretable concept structure in vocabulary space \citep{geva2022transformer}. Because different readout methods may produce scores with different offsets and scales, as also observed when comparing logit distributions across sources \citep{sun2024logit}, we standardize each readout with respect to its vocabulary distribution at the current decoding step: $\hat z_{\ell,t}=
\frac{z_{\ell,t}-\text{Mean}\left(z_{\ell,t}\right)}{\text{Std}\left(z_{\ell,t}\right)},
\label{eq:standardized_readout}$ where \(\operatorname{Mean}(\cdot)\) and \(\operatorname{Std}(\cdot)\) are computed over the vocabulary dimension. This places candidate readouts on a common scale without requiring learned normalization parameters or readout-specific calibration. Motivated by the bag-of-words attribute model of PPLM
\citep{dathathri2020plug}, which aggregates output-distribution probability mass over a curated word set to score an attribute, we introduce the \emph{value-specific readout score} $(s_{k,\ell,t})$, which aggregates standardized vocabulary evidence for value $V_k$ at an arbitrary intermediate layer $\ell$ rather than only at the final output distribution, with larger values indicating stronger value-specific evidence.

\begin{definition}[Value-Specific Readout Score]
\label{def:value_readout_score}
Given a fixed token vocabulary $(\mathcal{T}_k)$ for a targeted value \(V_k\), layer \(\ell\), and generation step \(t\), the \textit{value-specific readout score} is defined as
\begin{equation}
\small
\text{score}_{\ell,t}(V_k):=s_{k,\ell,t}
=
\frac{1}{|\mathcal T_k|}
\sum_{w\in\mathcal T_k}
\hat z_{\ell,t}[w].
\label{eq:raw_value_score}
\end{equation}
\end{definition}

We map this unbounded score to a bounded value state via a sigmoid,
following the use of sigmoid-transformed linear scores as bounded
confidence signals for gating or scaling activation interventions in
prior work \citep{hoscilowicz2024nliti, nguyen2025matsteer}: $c_{k,\ell,t} = \sigma\!\left(s_{k,\ell,t}\right) = \frac{1}{
1+\exp\left(-s_{k,\ell,t}\right) }. \label{eq:normalized_value_score}$ Thus, \(c_{k,\ell,t}\in(0,1)\), with larger values indicating stronger
observed expression of \(V_k\).

For layer \(\ell\) and generation step \(t\), the \emph{intermediate-layer value observer} maps the current intermediate
representation \(h_{\ell,t}\) to $\mathbf c_{\ell,t}=[c_{1,\ell,t},\ldots,c_{K,\ell,t}]^\top\in(0,1)^K. \label{eq:observer_state}$ The vector \(\mathbf c_{\ell,t}\) provides a bounded estimate of the
model's current value orientation under a readout. Unlike the
trained per-attribute gates used in prior sigmoid-based interventions
\citep{hoscilowicz2024nliti, nguyen2025matsteer}, our observer requires
no probe training: \(\mathbf c_{\ell,t}\) is computed directly from the
standardized vocabulary evidence in ~\eqref{eq:raw_value_score}, and
depends only on the current intermediate representation, using no
information from previous decoding steps. This keeps the feedback
mechanism lightweight, training-free, and responsive to changes during
generation.
\subsection{Observer-Guided Causal Multi-Value Steering}
\label{sec:controller}

This final stage of \texttt{AIMES} performs adaptive causal multi-value steering by
directly intervening on the model's hidden representations. We use \emph{causal} here in the interventionist sense; rather than only measuring associations in representation space, \texttt{AIMES} modifies the hidden state along value-specific directions and evaluates the resulting change in model behavior. For a fixed readout method and intervention layer \(\ell\), the observer
provides the current value state $\mathbf c_{\ell,t} = [c_{1,\ell,t},\ldots,c_{K,\ell,t}]^\top$ . Let $\mathbf g=[g_1,\ldots,g_K]^\top$, and
$g_k\in\{-1,0,+1\}$ encode the steering objective for value \(V_k\): \(+1\) for amplification,
\(-1\) for suppression, and \(0\) for no intervention. At each decoding step,
\texttt{AIMES} uses the current observer state to determine the contribution of
each requested value direction. The resulting joint intervention is
\begin{equation}
\small
h'_{\ell,t}
=
h_{\ell,t}
+
\gamma
\left[
\sum_{k:g_k=+1}
(1-c_{k,\ell,t})\,u_{k,\ell}
-
\sum_{k:g_k=-1}
c_{k,\ell,t}\,u_{k,\ell}
\right].
\label{eq:aimes_intervention}
\end{equation}

Here, \(\gamma>0\) is the shared base steering magnitude and
\(u_{k,\ell}\) is the unit-normalized direction for value \(V_k\).
The observer state \(c_{k,\ell,t}\) determines the value-specific steering
strength: amplification is weighted by $(1-c_{k,\ell,t})$, whereas suppression
is weighted by \(c_{k,\ell,t}\). The intervention therefore adapts to the
currently observed value state at each decoding step. The controller is
memoryless and depends only on the current readout, unlike PID-style
controllers that accumulate error over time
\citep{bharadwaj2025stupid}. Together, Sections~\ref{sec:value_directions}-\ref{sec:controller} define the complete \texttt{AIMES} framework: value directions specify the intervention axes, the intermediate-layer readout estimates the current value state, and the controller maps these estimates to adaptive steering coefficients during generation. The full \texttt{AIMES} algorithm and inference procedure are provided in Appendix~\ref{app:aimes_algorithm}.

\section{Experimental Analysis and Results}
\label{sec:experiments}
\paragraph{Experimental Setup and Baselines.}
We evaluate five instruction-tuned models spanning three families and parameter
scales: \texttt{Gemma-3-4B-IT} and \texttt{Gemma-3-12B-IT}
\citep{gemmateam2025gemma3}, \texttt{Qwen3-4B} and \texttt{Qwen3-14B}
\citep{yang2025qwen3}, and \texttt{Llama-3.1-8B-Instruct}
\citep{grattafiori2024llama3}. Following \citep{yu2026tracing}, we study the five MFT dimensions: \textit{Care, Fairness, Loyalty, Authority}, and \textit{Sanctity} \citep{graham2013MFT}. For each foundation,
we construct 200 matched positive-negative vignette pairs to derive value-specific steering directions.
\\
Our multi-value evaluation considers three objectives:
$(\uparrow\mathrm{Care}\uparrow\mathrm{Fairness})$, $(\uparrow\mathrm{Loyalty}\uparrow\mathrm{Authority})$, and $(\uparrow\mathrm{Care}\uparrow\mathrm{Fairness} \downarrow\mathrm{Sanctity})$, and each objective is evaluated across ten
model-specific intervention depths spanning the network. We compare \texttt{AIMES} with two non-adaptive baselines: \textbf{\textit{(i) Fixed Multi-Value Steering}} which uses the same directions \(u_{k,\ell}\), prompts, intervention layers, generation settings, and base magnitude as
\texttt{AIMES}, but applies all requested directions simultaneously with fixed strength,
$\Delta h_{\ell,t}=\gamma\sum_{k:g_k\neq0}g_k u_{k,\ell}$, without intermediate-state feedback. This is the direct multi-value extension of contrastive activation addition \citep{rimsky-etal-2024-steering} and
corresponds to naive vector composition in multi-attribute steering
\citep{vanderweij2024extending}. \textbf{\textit{(ii) Intensity-Anchor Prompt Steering}} which expresses the same multi-value objective directly in the prompt \citep{kim2026valueflow}, without modifying model activations. Because it is independent of intervention depth, the same prompt-steered response is compared
with \texttt{AIMES} at each tested layer. Behavioral control is evaluated using blind value-specific judgments from two independent evaluators, \texttt{GPT-5.6-Sol} \citep{openai2026gpt56} and \texttt{Claude-Opus-4.8} \citep{anthropic2026opus48}, applied to the same generated responses. Our main analysis reports matched prompt-level interaction contrasts and prompt-level consistency under both judges, while the full depth-wise cross-judge results are provided in Appendix~\ref{app:external_judge}. The interaction contrasts isolate the effect of adding a new value constraint, whereas prompt-level consistency measures how broadly the resulting advantage holds across examples. Response quality is evaluated separately using the blind \texttt{GPT-5.6-Sol} judge; complete metric definitions are provided in Appendix~\ref{app:eval_metrics}. All experiments are implemented with PyTorch and Hugging Face Transformers and run on NVIDIA A100 GPUs with Google Colab Pro+. Further details are provided in Appendix~\ref{app:experimental}. The full
code is publicly available at \url{https://anonymous.4open.science/status/AIMES-9A6C}.
\begin{figure}[t]
    \centering
    \includegraphics[scale=0.29]{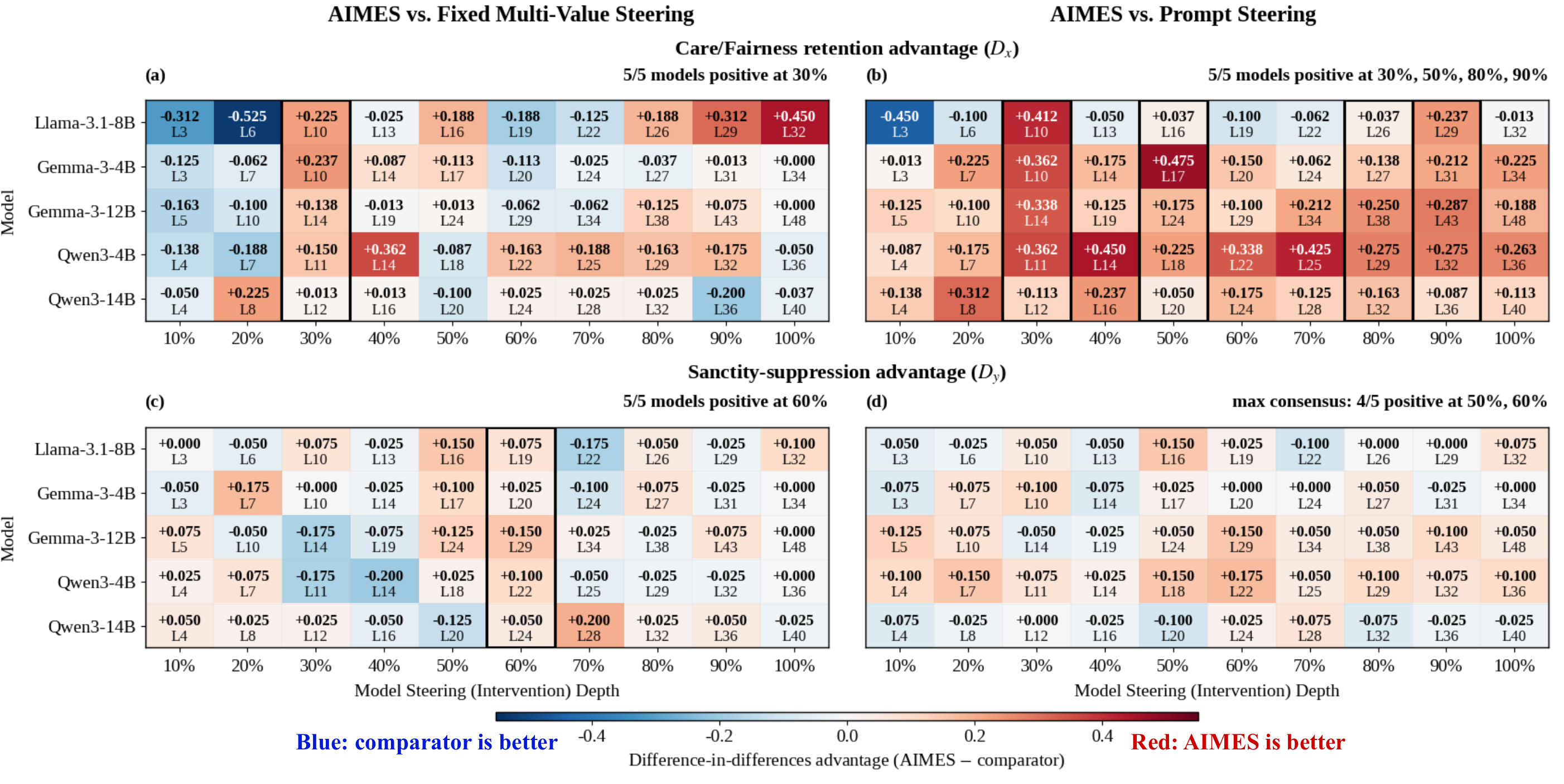}
    \caption{\textbf{Depth-wise \texttt{AIMES} advantage under GPT-5.6-Sol.} Panels~(a,b) show Care/Fairness retention (\(D_x\)) and panels~(c,d) Sanctity suppression (\(D_y\)) relative to Fixed Multi-Value Steering and Prompt Steering. Positive values favor \texttt{AIMES}; black outlines mark depths with unanimous positive advantage across all five models.}
    \label{fig:depth_consistency}
\end{figure}

\begin{table*}[t]
\centering
\setlength{\tabcolsep}{2.5pt}
\renewcommand{\arraystretch}{1.05}
\footnotesize
\begin{tabular}{llc cc cc cc cc}
\toprule
& & &
\multicolumn{4}{c}{\textbf{\texttt{AIMES} vs. Prompt Steering}} &
\multicolumn{4}{c}{\textbf{\texttt{AIMES} vs. Fixed Multi}} \\
\cmidrule(lr){4-7}\cmidrule(lr){8-11}

& & &
\multicolumn{2}{c}{\textbf{GPT-5.6-Sol}} &
\multicolumn{2}{c}{\textbf{Claude-Opus-4.8}} &
\multicolumn{2}{c}{\textbf{GPT-5.6-Sol}} &
\multicolumn{2}{c}{\textbf{Claude-Opus-4.8}} \\
\cmidrule(lr){4-5}\cmidrule(lr){6-7}
\cmidrule(lr){8-9}\cmidrule(lr){10-11}

\textbf{Model} &
\textbf{Stage} &
\textbf{Layer} &
$D_x$ & $D_y$ &
$D_x$ & $D_y$ &
$D_x$ & $D_y$ &
$D_x$ & $D_y$ \\

\midrule

\multirow{2}{*}{Llama-3.1-8B}
& Early
& 10
& \textbf{55.0} & \textbf{51.2}
& 48.8 & \textbf{55.0}
& \textbf{60.0} & \textbf{52.5}
& \textbf{55.0} & \textbf{51.2} \\

& Middle
& 19
& \textbf{53.8} & \textbf{52.5}
& 45.0 & 47.5
& 36.2 & \textbf{52.5}
& 48.8 & 45.0 \\

\midrule

\multirow{2}{*}{Gemma-3-4B}
& Early
& 10
& \textbf{67.5} & \textbf{52.5}
& \textbf{53.8} & 48.8
& \textbf{62.5} & 50.0
& \textbf{55.0} & 50.0 \\

& Middle
& 20
& \textbf{58.8} & \textbf{53.8}
& 50.0 & \textbf{51.2}
& 45.0 & 50.0
& 46.2 & \textbf{52.5} \\

\midrule

\multirow{2}{*}{Gemma-3-12B}
& Early
& 14
& \textbf{60.0} & \textbf{51.2}
& \textbf{55.0} & \textbf{55.0}
& 48.8 & 46.2
& 50.0 & 47.5 \\

& Middle
& 29
& \textbf{53.8} & \textbf{57.5}
& \textbf{57.5} & \textbf{58.8}
& 48.8 & \textbf{51.2}
& \textbf{57.5} & \textbf{53.8} \\

\midrule

\multirow{2}{*}{Qwen3-4B}
& Early
& 11
& \textbf{62.5} & \textbf{52.5}
& \textbf{58.8} & \textbf{51.2}
& \textbf{51.2} & 46.2
& \textbf{63.7} & 46.2 \\

& Middle
& 22
& \textbf{65.0} & \textbf{56.2}
& \textbf{51.2} & 50.0
& \textbf{60.0} & \textbf{55.0}
& \textbf{55.0} & 50.0 \\

\midrule

\multirow{2}{*}{Qwen3-14B}
& Early
& 12
& \textbf{51.2} & 50.0
& \textbf{56.2} & \textbf{55.0}
& 48.8 & \textbf{51.2}
& \textbf{55.0} & \textbf{51.2} \\

& Middle
& 24
& \textbf{56.2} & \textbf{51.2}
& \textbf{53.8} & \textbf{53.8}
& \textbf{51.2} & \textbf{51.2}
& 48.8 & 50.0 \\

\bottomrule
\end{tabular}

\caption{\textbf{Prompt-level consistency across judges.}
Tie-adjusted \texttt{AIMES} consistency (\%) at early ($30\%$) and middle ($60\%$) depths. $D_x$ measures Care/Fairness retention and $D_y$ Sanctity suppression.}
\label{tab:aimes_consistency}
\end{table*}
\paragraph{(1) Overall Analysis of Adaptive Multi-Value Steering.}
Figure~\ref{fig:depth_consistency} shows that the advantage of adaptive
steering depends on both intervention depth and the controlled objective.
Relative to Fixed Multi-Value Steering, \(D_x>0\) for all five models at
around \(30\%\) depth under \texttt{GPT-5.6-Sol}, a pattern that is analyzed
under the independent \texttt{Claude-Opus-4.8} evaluator (Appendix~\ref{app:external_judge}, Figure \ref{fig:claude_depth_consistency}). Under \texttt{GPT-5.6-Sol},
\(D_y>0\) for all five models at around \(60\%\) depth; however, the depth at
which this Sanctity-suppression advantage concentrates shifts under Claude,
indicating evaluator sensitivity. Against Prompt Steering, the Care/Fairness
advantage is more persistent under \texttt{GPT-5.6-Sol}, with \(D_x>0\) for
all five models at approximately \(30\%\), \(50\%\), \(80\%\), and \(90\%\)
depth, while the strongest \(D_y\) consensus reaches four of five models near
\(50\%\). Overall, the depth localization of the Care/Fairness-retention
advantage is more robust across evaluators than that of the
Sanctity-suppression advantage.
\paragraph{(2) Analyzing Prompt-Level Consistency.}
Table~\ref{tab:aimes_consistency} complements the mean interaction contrasts
in Figure~\ref{fig:depth_consistency} by measuring how broadly each advantage
is distributed across prompts. Against Prompt Steering, GPT-5.6-Sol yields
majority \(D_x\) consistency for all five models at both displayed depths,
whereas the Claude results are more heterogeneous. Relative to Fixed
Multi-Value Steering, consistency varies more strongly across models, depths,
and evaluators. Thus, the mean interaction contrasts characterize the average advantage, while prompt-level consistency indicates how broadly that advantage holds across examples. Full results across all ten depths are provided in
Appendix~\ref{app:incremental_robustness}.
\begin{figure}
    \centering
    \includegraphics[scale=0.22]{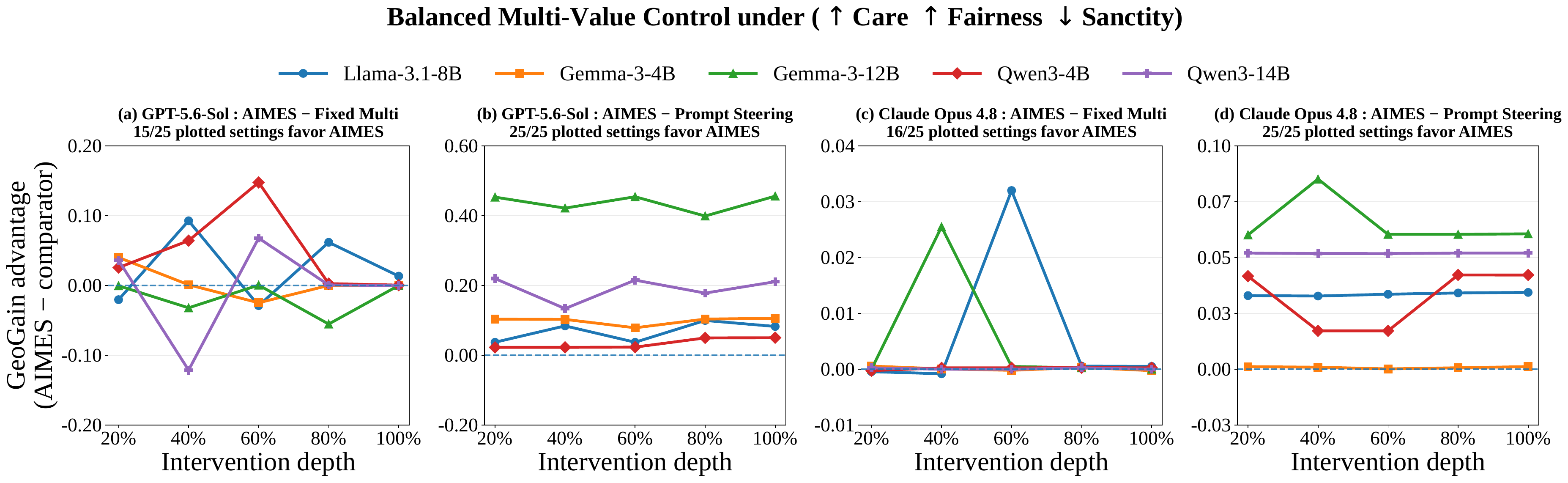}
    \caption{\textbf{GeoGain-based multi-value control evaluation.} GeoGain advantage of \texttt{AIMES} over Fixed Multi-Value Steering and Prompt Steering across sampled intervention depths, evaluated using two independent judges: GPT-5.6-Sol (a,b) and Claude-Opus-4.8 (c,d). In the figures, positive values favor \texttt{AIMES}; GeoGain measures balanced movement in $[\Delta C,\Delta F,-\Delta S]$.}
    \label{fig:GeoGain}
\end{figure}
\begin{figure}
    \centering
    \includegraphics[scale=0.3]{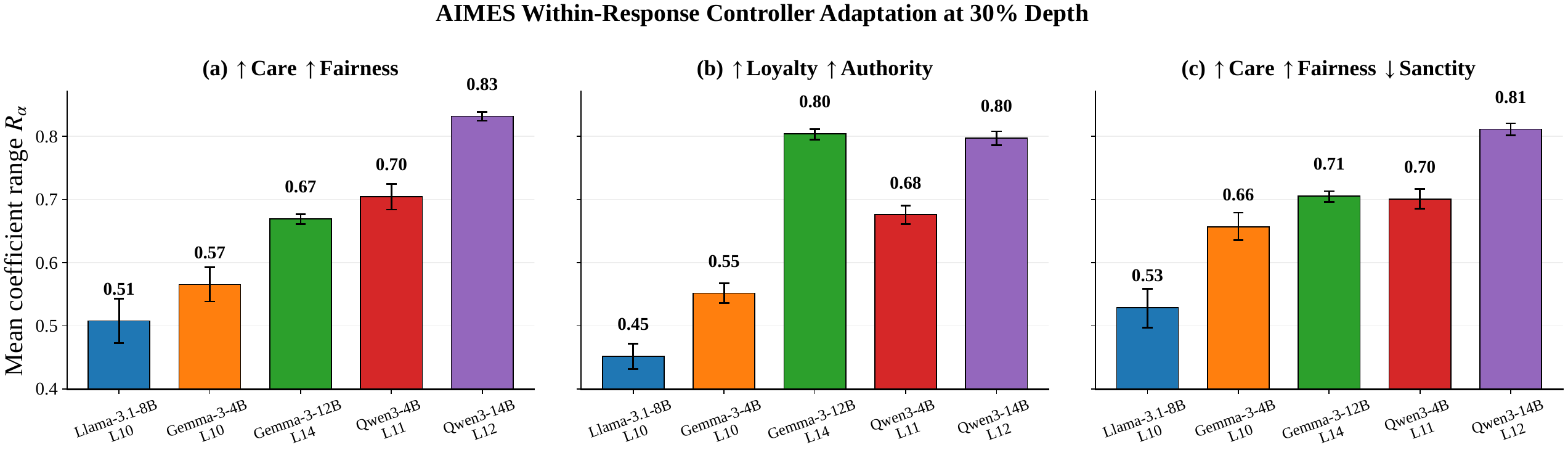}
    \caption{\textbf{Variation in AIMES steering strength within a response at 30\% depth.}
For each model and objective, we measure the within-response range of the steering coefficient, \(R_{\alpha,k}=\max_t\alpha_{k,t}-\min_t\alpha_{k,t}\). Larger values indicate greater changes in steering strength during generation. Bars show response means and error bars denote $95\%$ prompt-bootstrap confidence intervals.}
    \label{fig:controller_adaptation}
\end{figure}
\paragraph{(3) Evaluating Balanced Multi-Value Control.}
Figure~\ref{fig:GeoGain} evaluates joint control of
\((\uparrow\mathrm{Care}\uparrow\mathrm{Fairness}
\downarrow\mathrm{Sanctity})\) using GeoGain
\citep{ghasemi2026orbit}. Relative to Fixed Multi-Value Steering, the
advantage of \texttt{AIMES} remains model- and depth-dependent, with positive
GeoGain differences in 15/25 plotted settings under GPT-5.6-Sol and 16/25
under Claude-Opus-4.8. In contrast, \texttt{AIMES} achieves higher GeoGain than Prompt Steering in all 25 plotted settings under both judges. Thus, the balanced-control advantage is consistent relative to prompt-based steering, while the comparison with fixed joint steering remains dependent on model and
intervention depth. Full depth-wise results are provided in Appendix~\ref{app:results} (Figure~\ref{fig:GeoGain_full} and Figure \ref{fig:claude_geogain}).

\paragraph{(4) Within-Response Steering Variation.}
Compared to fixed activation steering, \texttt{AIMES} updates its intervention coefficients during generation using the online value readout. We measure this variation with the coefficient range
$(R_{\alpha,k}=\max_t\alpha_{k,t}-\min_t\alpha_{k,t})$.
Figure~\ref{fig:controller_adaptation} shows the mean range at the representative $30\%$ depth. Across all five models and three objectives, the ranges are consistently nonzero ($0.45$-$0.83$), showing that the steering strength varies within
responses rather than remaining fixed. Full results across all ten sampled depths are provided in Appendix~\ref{app:controller_adaptation}.

\section{Conclusion}

We introduced \texttt{AIMES}, a closed-loop framework for adaptive
activation-level steering of multiple human values using intermediate-layer
vocabulary readouts as online observers. Across five models from three
families, we find that multi-value controllability depends strongly on both
the steering objective and intervention depth. \texttt{AIMES} provides
depth-dependent advantages over fixed joint steering and, at favorable
depths, improves joint-control performance relative to prompt-based steering,
including for objectives that simultaneously promote and suppress values.
These gains are achieved with smaller realized activation-space interventions
than Fixed Multi-Value Steering while maintaining comparable response quality.
Within the Moral Foundations Theory values and model scales studied here,
\texttt{AIMES} offers a lightweight approach to adaptive multi-value control and motivates future work on broader value spaces, richer
observers, and geometry-aware controllers.

\subsection*{AI Use Statement}
Generative AI tools were used to assist with synthetic contrastive-vignette
generation, code debugging, and manuscript polishing and editing. All code, analyses, experimental decisions, and manuscript content were reviewed and verified by the authors. Synthetic data generated with AI tools were subjected
to the validation and filtering procedures described in Appendix~\ref{app:contrastive_generation}. The authors take full responsibility for the final content of the paper, including all reported results, claims, and artifacts produced with the aid of generative AI.

\subsection*{Ethics Statement}
This work studies inference-time control of value-related behavior in language
models. While such methods may support more adaptable model behavior, they
could also be used to impose particular value specifications or manipulate
model outputs in undesirable ways. Our experiments focus on the five Moral
Foundations Theory (MFT) dimensions and should not be interpreted as providing a
complete or universal representation of human values. The study uses synthetic
and publicly available evaluation data and does not involve direct human-subject experimentation.

\subsection*{Reproducibility Statement}

We provide the value-direction construction, observer definition, adaptive
controller, evaluation metrics, and experimental setup in the main paper.
The appendix further documents the contrastive-vignette generation procedure,
model-specific intervention layers, complete inference algorithm, observer
ablations, value-direction geometry, and full layer-wise results. An anonymous
code repository containing the implementation and supporting artifacts is
linked in the experimental section.

\bibliography{ref}
\bibliographystyle{iclr2027_conference}

\newpage
\appendix
\section*{Appendix}
The Appendix is organized as follows:
\begin{itemize}
    \item \ref{app:prelim_background} Background and Preliminaries
    \item \ref{app:aimes_algorithm} \texttt{AIMES} Algorithm
    \item \ref{app:experimental} Detailed Experimental Setup
    \item \ref{app:results} Additional Experimental Results
\end{itemize}
\section{Background and Preliminaries}
\label{app:prelim_background}

\paragraph{Activation-Space Control and Direction Composition.}
Activation steering modifies model behavior at inference time by perturbing
hidden representations along directions associated with desired concepts or
behaviors \citep{turner2023steering,zou2023representation,li2023inference}.
For human values, \citep{yu2026tracing} construct layer-specific contrastive
directions for the five Moral Foundations Theory dimensions
\citep{graham2013MFT} and show that interventions along these directions can
shift foundation-relevant behavior. Their analysis also reveals cross-value
effects, indicating that moral-value directions are not independent.

This issue is closely related to the broader problem of composing multiple
activation directions. Methods such as MAT-Steer
\citep{nguyen2025matsteer}, ORBIT \citep{ghasemi2026orbit}, MSRS
\citep{jiang2025msrs}, and K-Steering \citep{oozeer2025ksteering} address
interference among jointly controlled attributes using mechanisms including
gating, orthogonalization, subspace decomposition, rotation, and nonlinear
direction construction. These methods build on evidence that naive vector
addition can become unreliable when several attributes are controlled
simultaneously \citep{vanderweij2024extending}. In most cases, however, the
resulting intervention rule is determined before generation and remains fixed
during decoding.

\paragraph{Multi-Value Control.}
Recent work considers multiple human values through several complementary
mechanisms. VALUEFLOW \citep{kim2026valueflow} uses natural-language value
specifications to study how multiple value intensities interact, showing that
their effects need not compose additively. VISPA \citep{zheng2026vispa}
combines separately generated value-conditioned outputs for pluralistic
alignment, while \citep{yu2026tracing} intervene on moral-foundation
directions individually. Concurrently, \citep{pan2026spillover} study joint
activation-level multi-value steering and account for direction interference
through a static geometry-based correction.

These works highlight two distinct challenges in multi-value control:
\emph{geometric interference} among value directions and \emph{state
dependence} during generation. The former concerns how several directions
interact when applied jointly; the latter concerns whether their relative
strengths should remain fixed as the model's internal state evolves.
\texttt{AIMES} focuses on the second problem, using online feedback to adapt
the contributions of jointly applied value directions during decoding.

\paragraph{Feedback-Based Activation Steering.}
A separate line of work treats activation steering as a feedback-control
problem. \citep{nguyen2025feedback} distinguish fixed-strength open-loop
steering from controllers that adjust intervention magnitude according to the
difference between desired and observed concept states.
\citep{bharadwaj2025stupid} use PID-style feedback with a chunk-level
classifier during LLM reasoning, while \citep{prokopiou2026closing} study
feedback-controlled multi-attribute steering in symbolic music generation.
\citep{skifstad2026local} instead approximate layer-wise computation as a
locally linear dynamical system and derive interventions through
linear-quadratic regulation.

These approaches illustrate that intervention strength can be adjusted online,
but their feedback signals are typically obtained from a trained classifier,
probe, or fitted dynamical model. Our setting instead uses an
intermediate-layer vocabulary readout directly as the feedback signal,
allowing value-specific intervention strengths to be updated without training
a separate value-state estimator.

\paragraph{Layer-Wise Vocabulary Readouts.} \label{app:readouts}
Several methods map intermediate transformer representations into vocabulary space to make hidden states more interpretable. The \emph{Logit Lens} \citep{nostalgebraist2020logitlens} directly applies the model's unembedding matrix to an intermediate residual-stream activation

\begin{align}
z_{\ell,t}^{\mathrm{LL}} = W_U\,\mathrm{norm}(h_{\ell,t}),
\end{align}

effectively treating the intermediate state as if it were already expressed in final-layer output coordinates. The \emph{Tuned Lens} \citep{belrose2023tuned} instead learns a layer-specific affine transformation, 

\begin{align}
z_{\ell,t}^{\mathrm{TL}} =  W_U\,\mathrm{norm} \left( A_{\ell}h_{\ell,t}+b_{\ell} \right),
\end{align}

to better align intermediate activations with the model's final output space. J-Lens \citep{gurnee2026jlens} takes a different approach, using a layer-specific average Jacobian \(J_\ell\in\mathbb{R}^{d\times d}\) to approximate how information at layer \(\ell\) propagates toward the final representation:
\begin{align}
    z_{\ell,t}^{J}
    =
    W_U\,\mathrm{norm}(J_\ell h_{\ell,t}).
\end{align}
The corresponding token directions form the layer-specific dictionary
\(D_\ell=(W_UJ_\ell)^\top\in\mathbb{R}^{d\times|\mathcal V|}\), providing a vocabulary-aligned view of the residual stream.

Beyond interpretability, intermediate-layer vocabulary readouts have recently
been connected to activation steering. \citep{billa2026predicting} show that
Logit-Lens-based accessibility can help predict both concept steerability and
effective intervention depth across models and concept families. However,
representational visibility does not necessarily coincide with steerability:
\citep{nadaf2026steerable} show that effective steering directions can remain
weakly exposed or absent under the Logit Lens. This distinction is particularly
relevant to \texttt{AIMES}, where the readout is used not only to inspect the
intermediate representation, but also as the feedback signal that determines
the strength of the subsequent intervention. We therefore compare candidate readouts as online observers under the same steering setup in
Appendix~\ref{app:observer_ablation}.

We briefly review the two lines of work most closely related to our setting: J-lens representations and moral-value tracing in LLM activation space. J-Lens \citep{gurnee2026jlens} maps intermediate residual-stream activations
into vocabulary-aligned coordinates while accounting for how information
propagates through the remaining layers of the model. Let
$h_{\ell,t}\in\mathbb{R}^d$ denote the residual-stream activation at layer
$\ell$ and token position $t$, and let
$W_U\in\mathbb{R}^{|\mathcal{V}|\times d}$ denote the unembedding matrix.
J-Lens introduces a layer-specific average Jacobian
\begin{align}
    J_{\ell}
    =
    \mathbb{E}_{\mathrm{prompt},\,t,\,t'\geq t}
    \left[
        \frac{\partial h_{L,t'}}{\partial h_{\ell,t}}
    \right],
\end{align}
which provides a first-order approximation of how an intermediate
representation influences later final-layer representations. The corresponding
vocabulary-aligned readout is
\begin{align}
    z_{\ell,t}^{J}
    =
    W_U\,\mathrm{norm}(J_\ell h_{\ell,t}).
\end{align}
\texttt{AIMES} uses this intermediate vocabulary-space signal to construct the
value-specific observer scores defined in Section~\ref{sec:observer}.

\paragraph{Moral-Value Representations in LLMs.}
\citep{yu2026tracing} study how the five Moral Foundations Theory dimensions: \textit{Care, Fairness, Loyalty, Authority}, and \textit{Sanctity} \citep{graham2013MFT} are represented in LLM residual streams and whether
these representations can be behaviorally influenced through intervention.
For a target foundation \(V_k\), they construct a layer-specific direction
from the difference between mean activations of a positive set
\(\mathcal{E}_k^{+}\) and a contrast set \(\mathcal{E}_k^{-}\):
\begin{align}
    u_{k,\ell}^{\mathrm{raw}}
    =
    \frac{1}{|\mathcal{E}_k^{+}|}
    \sum_{i\in\mathcal{E}_k^{+}}
    \tilde{h}_{\ell}^{(i)}
    -
    \frac{1}{|\mathcal{E}_k^{-}|}
    \sum_{j\in\mathcal{E}_k^{-}}
    \tilde{h}_{\ell}^{(j)}, 
\end{align}
and normalizing that, we obtain:
\begin{align}
u_{k,\ell}=\frac{u_{k,\ell}^{\mathrm{raw}}}
    {\|u_{k,\ell}^{\mathrm{raw}}\|_2}
\end{align}
where \(\tilde{h}_{\ell}^{(i)}\) denotes the residual-stream representation
at the final input-token position. The resulting
\(u_{k,\ell}\in\mathbb{R}^{d}\) defines a layer-specific axis associated with
the target moral foundation.

The authors further analyze these directions using pretrained sparse
autoencoders (SAEs), identifying decoder features aligned with
\(u_{k,\ell}\) and interpreting them through their top-activating contexts.
They then test the behavioral relevance of the learned directions through
inference-time interventions of the form
\begin{align}
    h'_{\ell,t}
    =
    h_{\ell,t}
    +
    \alpha_{\ell}u_{k,\ell}.
\end{align}
These results provide evidence that moral-foundation information is organized
along distinguishable layer-wise activation directions whose manipulation can
shift foundation-relevant behavior. \texttt{AIMES} builds on this
contrastive-direction framework, but uses matched positive-negative poles for
each foundation and extends the setting from single-value interventions to
adaptive joint multi-value control.

\section{\texttt{AIMES} Algorithm} \label{app:aimes_algorithm}
\paragraph{\texttt{AIMES}: Inference-Time Adaptive Multi-Value Steering.}
For each generation run, \texttt{AIMES} fixes an intervention layer \(\ell\), readout \(r\), and base steering magnitude \(\gamma\). The intervention layer is varied across experiments to study depth effects, but remains fixed within a response. Every \(N\) decoding steps, the selected readout estimates the current expression of each controlled value and updates the corresponding adaptive coefficients. We use $N{=}1$ in all main experiments; $N>1$ reduces readout cost. These coefficients determine the relative contribution of each requested value direction until the next controller update. 

We use a fixed value-specific token set $\mathcal{T}_k$ for each moral foundation, with ten surface-form tokens per value. The complete token sets are listed in Table~\ref{tab:value_tokens}. These sets are fixed before evaluation and used consistently across all models and steering objectives.

\begin{table}[h]
\centering
\small
\begin{tabular}{lp{0.72\linewidth}}
\toprule
\textbf{Value} & \textbf{Observer token set $\mathcal{T}_k$} \\
\midrule
Care &
care, compassion, kindness, empathy, mercy, protection, welfare, nurture, support, helping \\
Fairness &
fairness, justice, equality, equity, impartiality, reciprocity, rights, fair, equitable, unbiased \\
Loyalty &
loyalty, fidelity, solidarity, allegiance, commitment, devotion, unity, belonging, fellowship, dedication \\
Authority &
authority, respect, duty, order, obedience, hierarchy, leadership, discipline, legitimacy, tradition \\
Sanctity &
purity, sanctity, sacred, holiness, dignity, reverence, virtue, innocence, cleanliness, sacredness \\
\bottomrule
\end{tabular}
\caption{\textbf{Value-specific observer token sets.}
Each set $\mathcal{T}_k$ contains ten fixed surface-form tokens associated with the corresponding Moral Foundations Theory value and is used to compute the value-specific readout score.}
\label{tab:value_tokens}
\end{table}
Because each coefficient is computed from the observed state of its corresponding value while all requested directions are applied jointly, the
same mechanism naturally supports both single-value and multi-value objectives. In our multi-value experiments, \texttt{AIMES} is compared with Fixed
Multi-Value Steering, which uses the same value directions, intervention layers, generation settings, and base magnitude \(\gamma\) but fixed
coefficients, and with Intensity-Anchor Prompt Steering, which expresses the
same objective through the input prompt without modifying model activations.

The complete \texttt{AIMES} framework is summarized in Algorithm \ref{alg:aimes}.

\begin{algorithm}[h]
\caption{\texttt{AIMES}:\textbf{A}daptive \textbf{I}ntervention for \textbf{M}ulti-Value \textbf{E}valuation and \textbf{S}teering}
\label{alg:aimes}
\begin{algorithmic}[1]
\Require Model \(M\), prompt \(x_i\), intervention layer \(\ell\),
value directions \(\{u_{k,\ell}\}_{k=1}^{K}\),
steering objective \(\mathbf g\),
token sets \(\{\mathcal T_k\}_{k=1}^{K}\),
readout \(r\), base steering magnitude \(\gamma\)
\Ensure Generated response \(y_i\)

\For{\(t=1,\ldots,T_i\)}
    \State Compute the pre-intervention hidden state \(h_{\ell,t}\)
    \State Compute \(z_{\ell,t} \gets \text{readout}(h_{\ell,t})\)
    \State Standardize \(z_{\ell,t}\) as $\hat z_{\ell,t}=
\frac{z_{\ell,t}-\text{Mean}\left(z_{\ell,t}\right)}{\text{Std}\left(z_{\ell,t}\right)}$  (highlighted in Section \ref{sec:observer})

    \For{each \(k\) such that \(g_k\neq0\)}
        \State Compute value $V_k$ specific score \(s_{k,\ell,t}\) using \eqref{eq:raw_value_score}
        \State Compute steering weights \(c_{k,\ell,t} \gets \sigma(s_{k,\ell,t})\)
    \EndFor

    \State Apply the joint adaptive intervention in
    \eqref{eq:aimes_intervention} to obtain steered state \(h'_{\ell,t}\)

    \State Continue the forward pass from \(h'_{\ell,t}\) and generate the next token
\EndFor

\State \Return \(y_i\)
\end{algorithmic}
\end{algorithm}
\newpage
\section{Detailed Experimental Setup}
\label{app:experimental}
\subsection{Construction and Overview}
\paragraph{Value-Direction Construction.}
\label{app:direction_construction} We study the five Moral Foundations Theory dimensions: Care, Fairness, Loyalty, Authority, and Sanctity. For each foundation, we construct $200$ matched positive-negative vignette pairs
corresponding to Care/Harm, Fairness/Cheating, Loyalty/Betrayal,
Authority/Subversion, and Sanctity/Degradation, and use them to derive value-specific steering directions. Each pair describes the same underlying setting while reversing the relevant moral orientation, reducing
variation due to topic, actors, and writing style.

For each model and transformer layer, we extract the final-token hidden
representation of the positive and negative examples and compute the
normalized contrastive mean-difference direction described in
Section~\ref{sec:value_directions}. Direction-construction examples are
kept disjoint from all prompts used for behavioral evaluation. The same
precomputed directions are used by Fixed
Multi-Value Steering and \texttt{AIMES}, ensuring that differences
between methods arise from the intervention strategy rather than from different direction estimates.

\paragraph{Multi-Value Evaluation Data.}
For the main multi-value experiments, we construct a separate evaluation set
of \(100\) held-out MFRC prompts, balanced across the five foundations
(\(20\) prompts per foundation). For each steering objective, we retain only
prompts whose source foundation is not one of the values being directly
controlled. This avoids evaluating a multi-value intervention on prompts that
are already explicitly associated with one of its target foundations.

Under this filtering, the two-value objectives
\((\uparrow\mathrm{Care}\uparrow\mathrm{Fairness})\) and
\((\uparrow\mathrm{Loyalty}\uparrow\mathrm{Authority})\) each retain
\(60\) prompts, corresponding to the three non-target foundations. The
three-value competing objective
\((\uparrow\mathrm{Care}\uparrow\mathrm{Fairness}
\downarrow\mathrm{Sanctity})\) retains \(40\) prompts from the two remaining
foundations, Loyalty and Authority. Each objective is evaluated at all ten
model-specific intervention depths using the same prompts for
\texttt{AIMES}, Fixed Multi-Value Steering, and Prompt Steering.

For the matched interaction analysis comparing
\((\uparrow\mathrm{Care}\uparrow\mathrm{Fairness}
\downarrow\mathrm{Sanctity})\) with
\((\uparrow\mathrm{Care}\uparrow\mathrm{Fairness})\), we restrict both
objectives to the same \(40\) prompts. Thus, the \(D_x\), \(D_y\), and
prompt-level consistency analyses use \(n=40\) matched prompts per
model-layer-comparator setting.
\paragraph{Intervention Layers.}
\label{app:intervention_layers} Since the evaluated architectures contain different numbers of
transformer layers, we compare intervention locations using normalized
depth
\[
d_\ell=\frac{\ell}{L},
\]
where \(L\) is the number of transformer layers. For each model, we
evaluate ten approximately evenly spaced depths spanning the network.
The exact layer indices are reported in Table~\ref{tab:intervention_layers}. The intervention layers are selected a priori to approximately span the
network depth; no layer is selected post hoc based on behavioral performance. The complete depth sweep is used to characterize how steering
behavior changes through the network.

\begin{table}[h]
\centering
\small
\begin{tabular}{lcc}
\toprule
\textbf{Model} & \textbf{No. Layers} & \textbf{Intervention Layers} \\
\midrule
Gemma-3-4B-IT
& 34
& $3,7,10,14,17,20,24,27,31,34$ \\

Gemma-3-12B-IT
& 48
& $5,10,14,19,24,29,34,38,43,48$ \\

Qwen3-4B
& 36
& $4,7,11,14,18,22,25,29,32,36$ \\

Qwen3-14B
& 40
& $4,8,12,16,20,24,28,32,36,40$ \\

Llama-3.1-8B-Instruct
& 32
& $3,6,10,13,16,19,22,26,29,32$ \\
\bottomrule
\end{tabular}
\caption{
Transformer layers used in the depth sweep. The layer indices are chosen
to approximately cover normalized depths from \(0.1\) to \(1.0\) for each
architecture.
}
\label{tab:intervention_layers}
\end{table}

\paragraph{Multi-Value Steering.}
\label{app:multivalue_setup}

The multi-value experiments compare \texttt{AIMES} with two non-adaptive
baselines: Fixed Multi-Value Steering and Intensity-Anchor Prompt Steering.
Fixed Multi-Value Steering jointly applies all requested value directions at the same
intervention layer using fixed coefficients, while \texttt{AIMES} uses the
same prompts, value directions, intervention layers, generation settings, and
base magnitude but replaces the fixed coefficients with the
observer-dependent weights defined in Section~\ref{sec:controller}. Prompt
Steering specifies the same multi-value objective directly in the input prompt
without modifying model activations. Unsteered responses are used as matched
references when computing value-score changes. For each \texttt{AIMES} generation, the intervention layer, observer, and base magnitude are fixed, while the value-specific coefficients are recomputed at every decoding step (\(N=1\) in all main experiments). The controller is memoryless and retains no accumulated state across decoding steps.

\paragraph{Metrics and Statistical Analysis.}
All method comparisons are performed on matched prompts. Our primary analysis
uses matched interaction contrasts to measure how adding a value constraint
changes control relative to a shared base objective, together with
prompt-level consistency to characterize how broadly the effect holds across
examples. We additionally use GeoGain \citep{ghasemi2026orbit} to evaluate
balanced movement across all requested values, and separately report blind
response quality and realized activation-space perturbation. Pointwise
bootstrap confidence intervals are used where reported. Complete metric
definitions and statistical details are provided in
Appendix~\ref{app:eval_metrics}.

\subsection{Contrastive Vignette Generation}
\begin{figure}[!ht]
    \centering
    \includegraphics[scale=0.26]{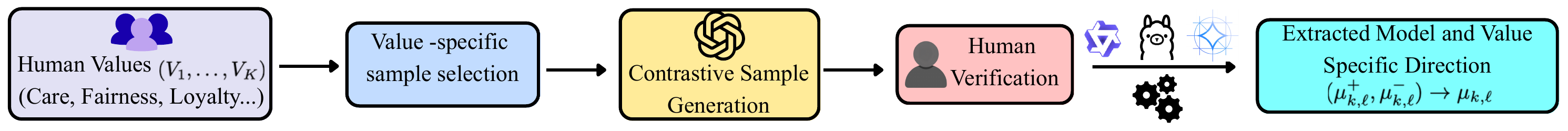}
    \caption{\textbf{Value-specific steering direction construction.} Value examples are selected, converted into contrastive pairs, verified, and used to extract model and value-specific activation directions.}
    \label{fig:value_dir_workflow}
\end{figure}
\paragraph{Independent semantic validity check.} We have generated the contrastive value-specific and model-specific directions as shown in Figure \ref{fig:value_dir_workflow}. 
As a preliminary sanity check, we evaluated whether each generated foundation-specific vignette set was semantically distinguishable from the Social Norm reference set using three independent sentence-embedding models: BGE-large-en-v1.5, E5-large-v2, and all-MPNet-base-v2. For foundation \(V_k\) and embedding model \(m\), we computed
\begin{align}
\Delta_k^{(m)}
=
\mathbb E_{x\sim\mathcal D_k}
\left[
\cos\!\left(e_m(x),e_m(d_k)\right)
\right]
-
\mathbb E_{x\sim\mathcal D_{\mathrm{SN}}}
\left[
\cos\!\left(e_m(x),e_m(d_k)\right)
\right],
\end{align}
where \(e_m(\cdot)\) denotes the representation produced by embedding model \(m\), and \(d_k\) denotes the textual definition of foundation \(V_k\). A positive \(\Delta_k^{(m)}\) therefore indicates that the foundation-specific vignette set is, on average, more semantically aligned with its intended foundation definition than the Social Norm reference set. All five foundations yielded positive contrasts under all three embedding models, indicating consistent set-level separation from the Social Norm distribution.

\begin{table}[t]
\centering
\begin{tabular}{lcc}
\toprule
\textbf{Foundation} & \(\boldsymbol{\overline{\Delta}_k}\) & \textbf{Positive Models} \\
\midrule
Care      & 0.063 & 3/3 \\
Fairness  & 0.052 & 3/3 \\
Loyalty   & 0.060 & 3/3 \\
Authority & 0.035 & 3/3 \\
Sanctity  & 0.078 & 3/3 \\
\bottomrule
\end{tabular}
\caption{Independent semantic sanity check of the generated foundation-specific vignette sets. \(\overline{\Delta}_k\) denotes the mean foundation-versus-Social-Norm cosine-similarity \citep{salton1975vector} difference averaged across BGE-large-en-v1.5, E5-large-v2, and all-MPNet-base-v2.}
\label{tab:semantic_sanity}
\end{table}

Averaged across embedding models, the observed differences were \(0.063\) for Care, \(0.052\) for Fairness, \(0.060\) for Loyalty, \(0.035\) for Authority, and \(0.078\) for Sanctity, with Authority showing the weakest separation and Sanctity the strongest. Because these embedding models are general-purpose semantic encoders rather than Moral Foundations Theory classifiers, we use this analysis only as a corpus-level sanity check. This foundation-versus-Social-Norm comparison also provides a semantic counterpart to the contrastive construction used in prior moral-tracing work \citep{yu2026tracing}, while the matched positive--negative corpus described separately is used to construct the bidirectional steering directions employed by \texttt{AIMES}. The primary validation of these directions is subsequently performed in the target models' activation space.

\paragraph{Matched Positive-Negative Contrastive Vignette Generation.}
\label{app:contrastive_generation} For the primary \texttt{AIMES} steering directions, we construct a second synthetic corpus consisting of matched positive--negative vignette pairs for each of the five Moral Foundations Theory dimensions. In contrast to the foundation-versus-Social-Norm corpus used for reproduction and representation-level comparison, this corpus is designed specifically to recover a bidirectional moral axis suitable for activation steering.

For each moral foundation $V_k$, we define the corresponding positive and negative poles according to the standard Moral Foundations Theory formulation:
\begin{align}
\mathrm{Care} &\leftrightarrow \mathrm{Harm},\\
\mathrm{Fairness} &\leftrightarrow \mathrm{Cheating},\\
\mathrm{Loyalty} &\leftrightarrow \mathrm{Betrayal},\\
\mathrm{Authority} &\leftrightarrow \mathrm{Subversion},\\
\mathrm{Sanctity} &\leftrightarrow \mathrm{Degradation}.
\end{align}
These paired dimensions follow the canonical virtue--vice structure of Moral Foundations Theory \citep{graham2011mapping,graham2013MFT}. For each foundation, we generate $N=200$ matched contrastive pairs,
\begin{align}
\mathcal D_k^{\pm}
=
\left\{
\left(
x_{k,i}^{+},
x_{k,i}^{-}
\right)
\right\}_{i=1}^{200},
\end{align}
where $x_{k,i}^{+}$ expresses the positive pole of the foundation and $x_{k,i}^{-}$ expresses its corresponding negative pole. Equivalently, these matched pairs induce positive and negative example sets
\begin{align}
\mathcal D_k^{+}
&=
\left\{
x_{k,i}^{+}
\right\}_{i=1}^{200},
\qquad
\mathcal D_k^{-}
=
\left\{
x_{k,i}^{-}
\right\}_{i=1}^{200}.
\end{align}
Across the five foundations, this produces $1{,}000$ matched pairs, or $2{,}000$ individual vignettes.

The two members of each pair are generated jointly rather than independently. The generator is instructed to preserve, as closely as possible, the same actors, setting, underlying event, grammatical structure, level of detail, and approximate length across the positive and negative versions. The principal difference between the two members should be the moral polarity associated with the target foundation. This matched construction reduces variation due to topic, setting, lexical content, and writing style, making the resulting activation difference more directly attributable to the target moral dimension.

For example, a Care/Harm pair may take the form
\begin{quote}
\textbf{Care:} A student stopped to comfort a classmate who was crying after being mocked.\\
\textbf{Harm:} A student joined the others in mocking a classmate who was already crying.
\end{quote}
Similarly, a Fairness/Cheating pair may contrast returning money accidentally overpaid by a customer with intentionally keeping the excess amount. The specific wording is not fixed; rather, the generation procedure enforces structural similarity within each pair while varying scenarios across the corpus.

To promote contextual diversity, the same twelve generation-time social settings used in the foundation-versus-Social-Norm corpus are retained:
\begin{enumerate}
    \item family and household,
    \item friends and interpersonal relationships,
    \item school and education,
    \item workplace,
    \item community and neighborhood,
    \item groups, clubs, and organizations,
    \item public and service interactions,
    \item online and digital interactions,
    \item sports and recreation,
    \item health and caregiving,
    \item travel and transportation, and
    \item social gatherings.
\end{enumerate}
These settings are used only as diversity controls and are not treated as moral labels or theoretically defined context categories. Generation is approximately balanced across the twelve contexts, yielding roughly $16$--$17$ pairs per context for each foundation. Each generated record stores the pair identifier, target foundation, diversity context, positive vignette, negative vignette, generator identifier, and generation metadata. The resulting data therefore have the conceptual structure
\begin{align}
\left(
\texttt{pair\_id},
\texttt{foundation},
\texttt{context},
\texttt{positive\_text},
\texttt{negative\_text}
\right).
\end{align}

To construct the layer-wise activation direction, we first compute the mean hidden representation of the positive and negative sets at layer $\ell$:
\begin{align}
\mu_{k,\ell}^{+}
&=
\frac{1}{N}
\sum_{i=1}^{N}
h_\ell(x_{k,i}^{+}),
\qquad
\mu_{k,\ell}^{-}
=
\frac{1}{N}
\sum_{i=1}^{N}
h_\ell(x_{k,i}^{-}).
\end{align}
We then define the normalized positive--negative direction as
\begin{align}
u_{k,\ell}^{\mathrm{PN}}
=
\frac{
\mu_{k,\ell}^{+}
-
\mu_{k,\ell}^{-}
}{
\left\|
\mu_{k,\ell}^{+}
-
\mu_{k,\ell}^{-}
\right\|_2
}.
\end{align}
Because the two sets consist of one-to-one matched pairs of equal size, the unnormalized mean difference is equivalently
\begin{align}
\mu_{k,\ell}^{+}
-
\mu_{k,\ell}^{-}
=
\frac{1}{N}
\sum_{i=1}^{N}
\left[
h_\ell(x_{k,i}^{+})
-
h_\ell(x_{k,i}^{-})
\right].
\end{align}
Thus, the mean-difference and pairwise-difference formulations yield the same direction. Here, $h_\ell(\cdot)$ denotes the hidden representation extracted from layer $\ell$ using a fixed token-selection rule. This construction follows the general contrastive mean-difference principle used in activation-steering methods \citep{rimsky-etal-2024-steering}.

The positive--negative corpus is kept separate from the foundation-versus-Social-Norm corpus. The latter is used primarily for reproducing and comparing against the representation structure reported in prior moral-tracing work, whereas the positive--negative corpus provides the primary bidirectional steering directions used by AIMES.
\subsection{Observer Ablation: J-Lens vs. Logit Lens}
\label{app:observer_ablation}
\begin{figure}
    \centering
    \includegraphics[scale=0.35]{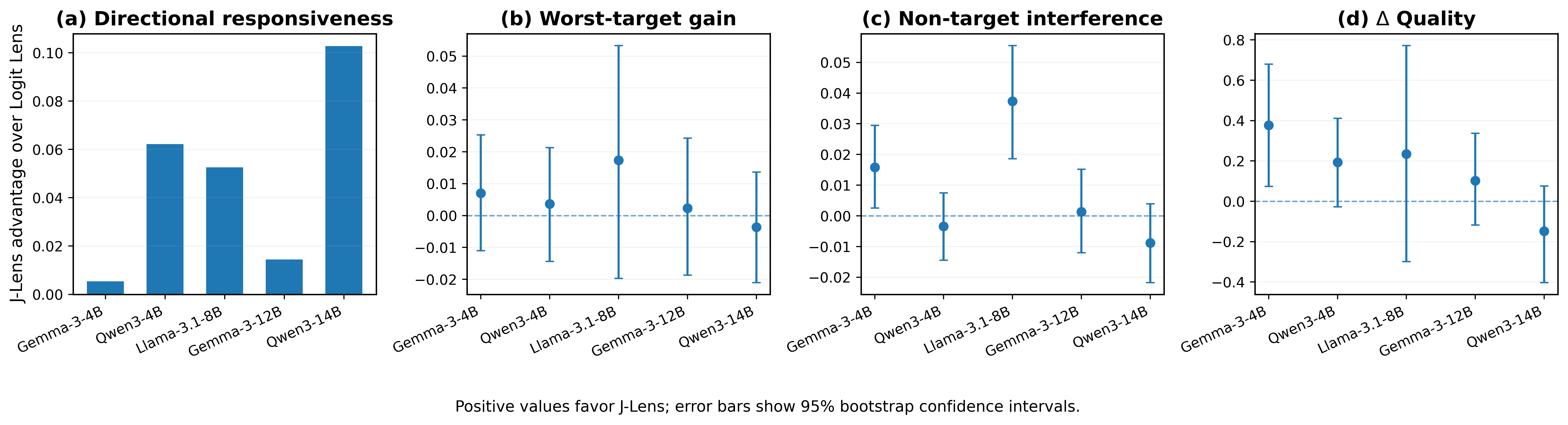}
    \caption{\textbf{Observer ablation: J-Lens versus Logit Lens.} Positive values favor J-Lens. J-Lens yields higher directional responsiveness in all five models and generally favorable downstream trends for worst-target gain, non-target interference, and response quality. Error bars in panels (b)-(d) denote $95\%$ bootstrap confidence intervals.}
    \label{fig:observer_ablation}
\end{figure}

Figure~\ref{fig:observer_ablation} compares J-Lens and Logit Lens as the
feedback observer while holding the steering directions, controller, intervention layers, prompts, and generation settings fixed. J-Lens
exhibits higher directional responsiveness in all five models, indicating
that its internal value estimates more consistently move in the requested
direction under intervention. Downstream, AIMES-JL achieves higher
worst-target gain in four of five models, lower non-target interference in
three of five, and higher response quality in four of five. Most behavioral
confidence intervals overlap zero, indicating that the downstream advantage
is not uniform across architectures; however, these results, together with
the consistently stronger directional responsiveness, motivate our use of
J-Lens as the default observer in the main experiments. 
\subsection{Layer-Wise Geometry of Value Directions}
\label{app:value_geometry}

\begin{figure}
    \centering
    \includegraphics[scale=0.3]{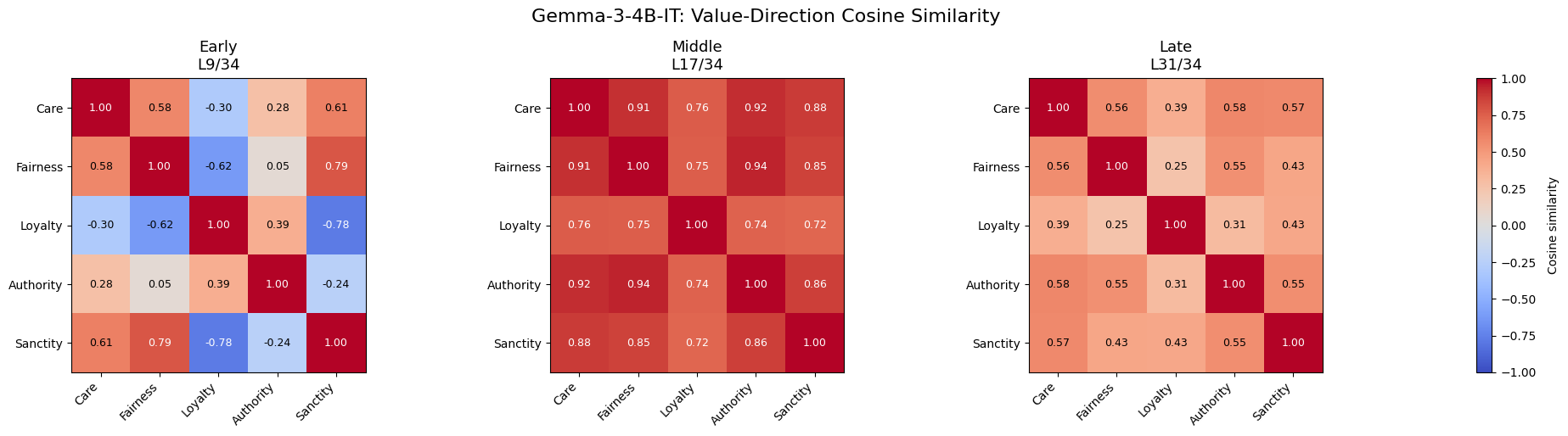}
    \includegraphics[scale=0.3]{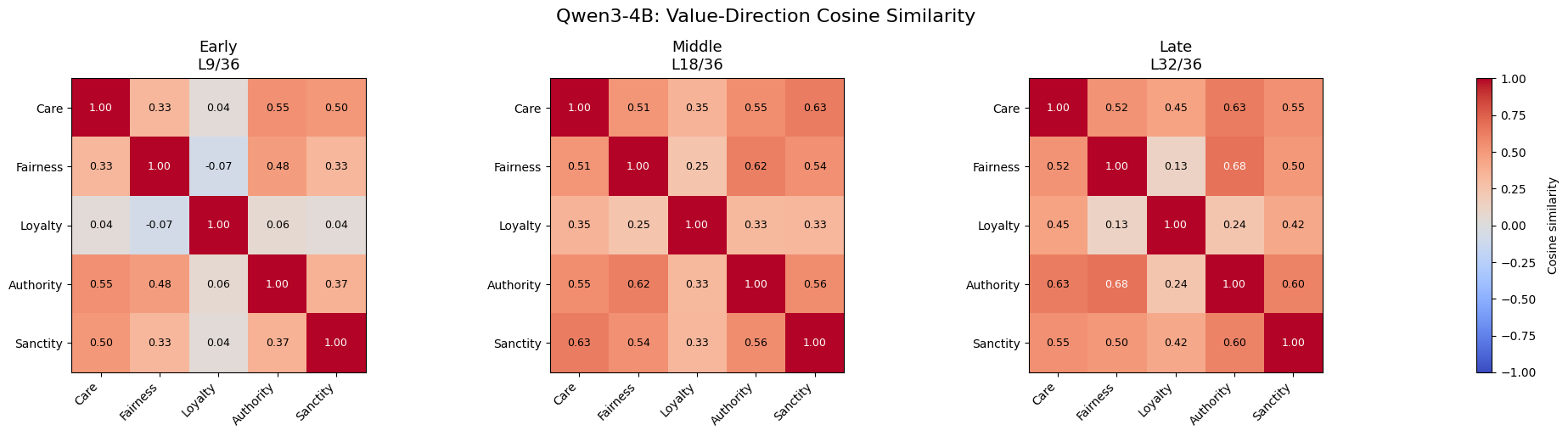}
    \includegraphics[scale=0.3]{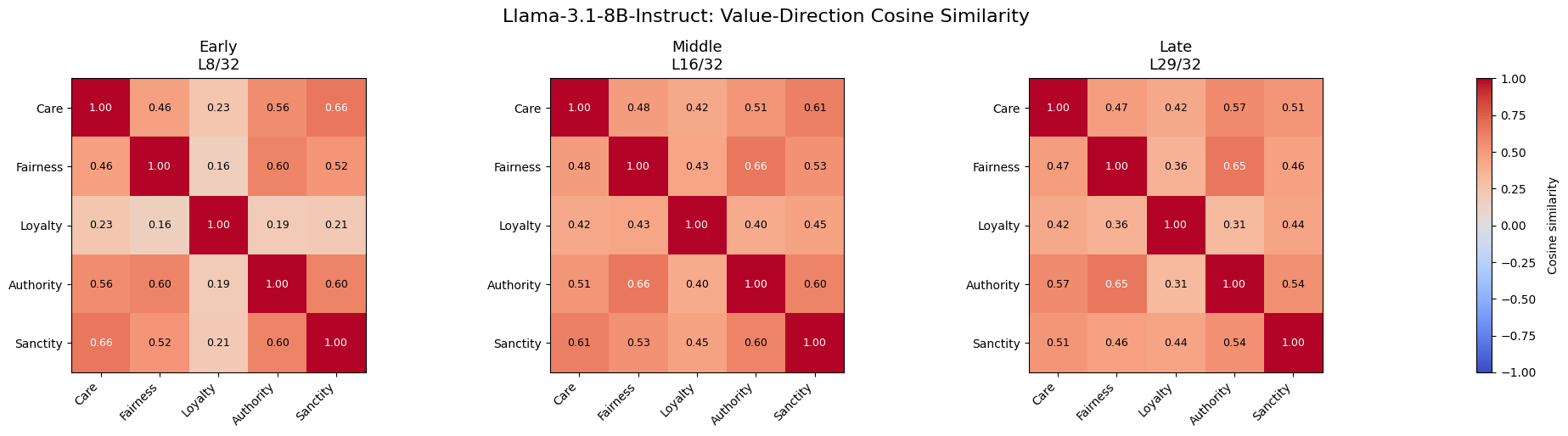}
    \includegraphics[scale=0.3]{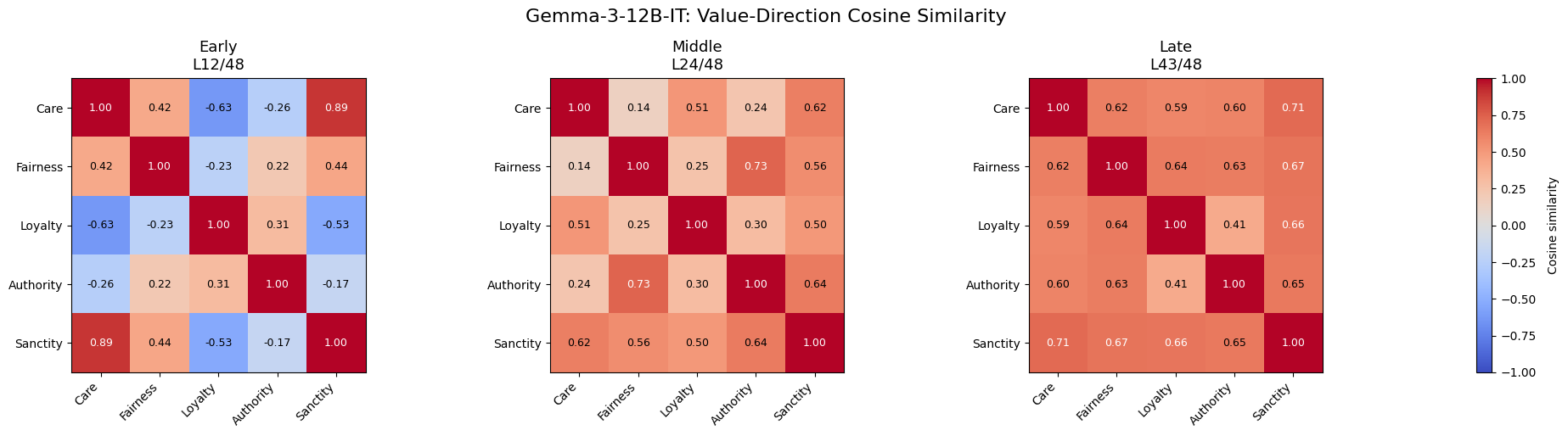}
    \includegraphics[scale=0.3]{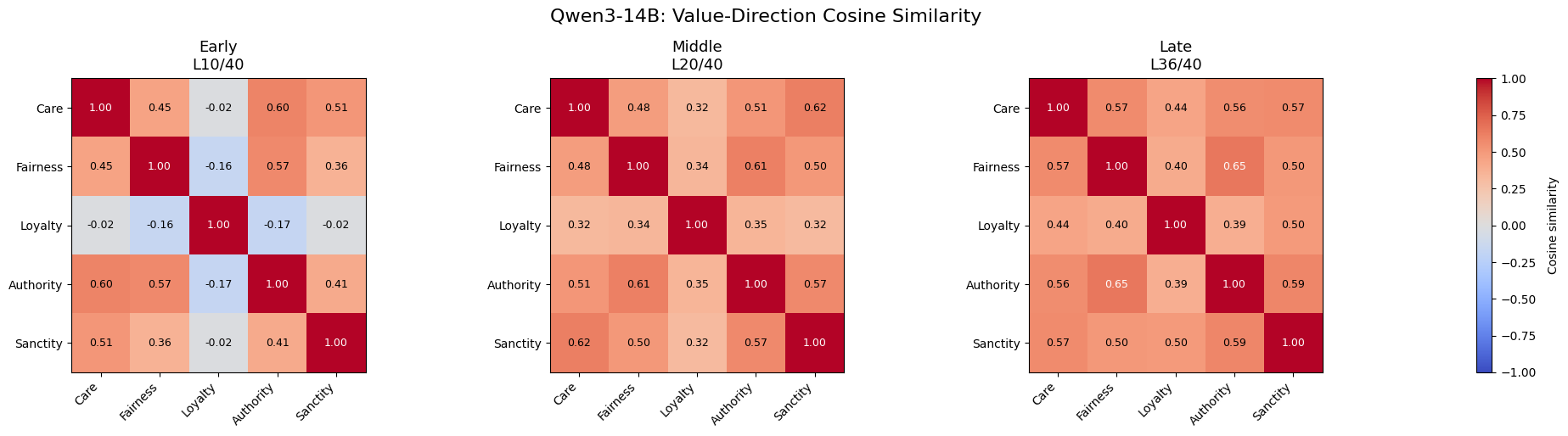}
    \caption{ Pairwise cosine similarity among the five value directions across models and network depth. Each model is shown at representative early, middle, and late layers, with entries reporting \(\cos(u_{i,\ell},u_{j,\ell})\) for Care, Fairness, Loyalty, Authority, and Sanctity. Positive values indicate aligned directions, values near zero indicate approximate orthogonality, and negative values indicate opposing geometry. }
    \label{fig:5model_3layer_cosine}
\end{figure}

Figure~\ref{fig:5model_3layer_cosine} shows the pairwise cosine similarity among the five value directions at representative early, middle, and late layers for all five models. The geometry is strongly model and depth-dependent. In Llama-3.1-8B-Instruct, pairwise similarities are predominantly positive and relatively stable across depth, although Loyalty remains less aligned with the other foundations. The two Qwen3 models exhibit a clearer depth-dependent transition: Loyalty is nearly orthogonal, and in some cases weakly anti-aligned, with several other directions at early layers, while the directions become more positively correlated in middle and late layers.

The Gemma models show the largest changes across depth. In Gemma-3-4B-IT, several early-layer pairs are negatively aligned, whereas the middle-layer directions become highly correlated before separating again toward later layers. Gemma-3-12B-IT similarly exhibits strong positive and negative relationships at early depth, followed by broadly positive correlations in the middle and late network. These results indicate that the interaction structure among value directions cannot be treated as fixed across either models or layers. Consequently, simultaneously applying several directions may produce substantially different geometric interactions depending on the intervention depth, motivating our evaluation of multi-value steering across multiple depths rather than selecting a single layer a priori.
\begin{figure}[h]
    \centering
    \includegraphics[scale=0.35]{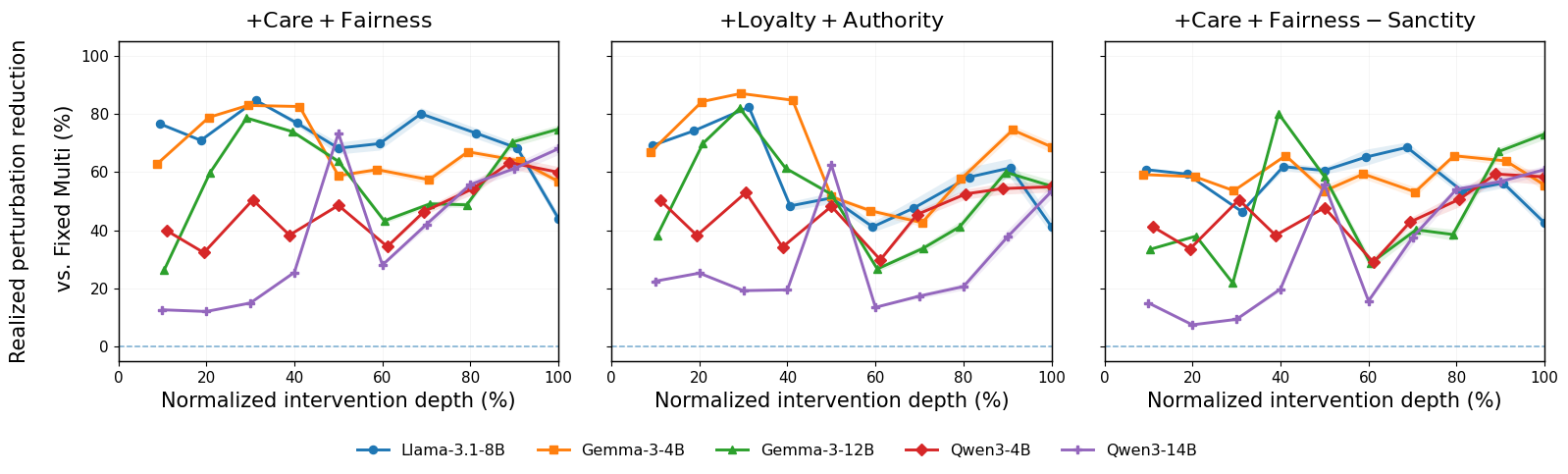}
    \caption{
\textbf{Realized activation-space perturbation reduction relative to Fixed
Multi-Value Steering}, $R_\ell = 100\left(1 - \|\Delta h_{\mathrm{AIMES},\ell}\|_2 /
\|\Delta h_{\mathrm{Fixed},\ell}\|_2\right)$, on shared prompt states.
$R_\ell>0$ in all $150$ model-layer-objective settings ($95\%$ CIs
excluding zero), with magnitude varying substantially by model, depth,
and objective rather than reflecting a constant rescaling.}
    \label{fig:realized_perturbation}
\end{figure}
\paragraph{\texttt{AIMES} achieves control with smaller realized interventions.}
Figure~\ref{fig:realized_perturbation} compares the realized activation-space
update of \texttt{AIMES} with Fixed Multi-Value Steering across all 150
model-layer-objective configurations. \texttt{AIMES} produces a smaller
perturbation in every setting, with reductions ranging from approximately
\(8\%\) to \(87\%\); the pointwise \(95\%\) bootstrap confidence interval
excludes zero throughout. The magnitude of the reduction varies substantially
across models, layers, and objectives, indicating that the adaptive controller
does not simply rescale the fixed intervention by a constant factor. Thus, the
behavioral gains reported in the main text are obtained with smaller realized
activation updates rather than stronger interventions.

\begin{figure}[h]
    \centering
    \includegraphics[scale=0.42]{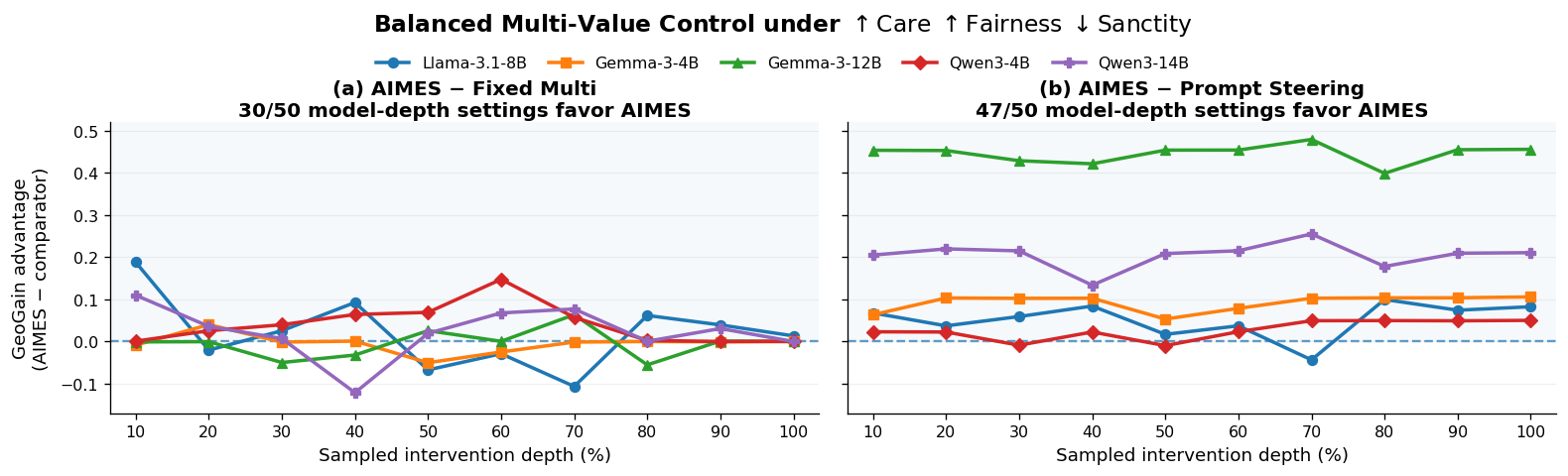}
    \caption{\textbf{GPT-5.6-Sol GeoGain advantage for balanced multi-value control under $(\uparrow\text{Care}\uparrow\text{Fairness}\downarrow\text{Sanctity})$.} GeoGain uses the geometric mean of the aligned target changes\([\Delta C,\Delta F,-\Delta S]\), with a positive sign only when all targets move in the requested directions. Curves show the prompt-paired mean (\texttt{AIMES}--comparator) difference across intervention depths; positive values favor \texttt{AIMES}.}
    \label{fig:GeoGain_full}
\end{figure}

\paragraph{Full Layer-Wise GeoGain Analysis.}
Figure~\ref{fig:GeoGain_full} extends the balanced-control analysis across the
full set of tested intervention depths. The comparison with Fixed Multi-Value Steering varies substantially across models and layers, confirming that the relative benefit of adaptive steering is depth-dependent. In contrast, the comparison with Prompt
Steering is more consistently positive across models and depth. These results
reinforce the main-text finding that balanced control under the competing
objective depends strongly on intervention location, particularly when
compared with fixed joint steering.

\section{Additional Experimental Results}\label{app:results}
\paragraph{Evaluation Metrics.}\label{app:eval_metrics}
We evaluate adaptive multi-value control using four complementary measures.
We use a shared base steering magnitude of \(\gamma=1.0\) for all main
multi-value experiments.
\paragraph{(a) Matched Interaction Contrast.} We use a matched interaction contrast, adapting the
difference-in-differences formulation
\citep{card1994minimum,angrist2009mostly}, to measure how adding a new value
constraint changes control relative to a shared base objective. Let
\(A=(\uparrow\mathrm{Care}\uparrow\mathrm{Fairness}
\downarrow\mathrm{Sanctity})\) and
\(B=(\uparrow\mathrm{Care}\uparrow\mathrm{Fairness})\). For prompt \(i\),
method \(m\), objective \(o\), and layer \(\ell\), we define
\begin{align}
G_{\mathrm{CF},i}^{(m,o,\ell)}
&=
\frac{\Delta C_i^{(m,o,\ell)}+\Delta F_i^{(m,o,\ell)}}{2},
&
G_{\mathrm{S},i}^{(m,o,\ell)}
&=
-\Delta S_i^{(m,o,\ell)},
\end{align}
where changes are measured relative to the corresponding unsteered response.
The incremental effect of adding \(\downarrow\mathrm{Sanctity}\) is
\(I_x=G_{\mathrm{CF}}^{A}-G_{\mathrm{CF}}^{B}\) for Care/Fairness retention
and \(I_y=G_{\mathrm{S}}^{A}-G_{\mathrm{S}}^{B}\) for additional Sanctity
suppression. For comparator
\(q\in\{\mathrm{Fixed},\mathrm{Prompt}\}\), we report
\begin{align}
D_x^{(q)}(\ell)
&=
\frac{1}{n}\sum_i
\left[
I_{x,i}^{(\mathrm{AIMES},\ell)}
-I_{x,i}^{(q,\ell)}
\right],
&
D_y^{(q)}(\ell)
&=
\frac{1}{n}\sum_i
\left[
I_{y,i}^{(\mathrm{AIMES},\ell)}
-I_{y,i}^{(q,\ell)}
\right].
\end{align}
Positive values favor \texttt{AIMES}. We additionally report
\emph{prompt-level consistency}, the tie-adjusted percentage of matched prompts
for which the corresponding (\texttt{AIMES}--comparator) incremental effect is positive.

\paragraph{(b) Geometric Gain (GeoGain).} To measure balanced progress across all components of a composite
objective, we use ORBIT-style Geometric Gain (GeoGain)
\citep{ghasemi2026orbit}. For aligned target changes
\(\tilde{\Delta}_k=g_k\Delta_k\), we compute
\[
M_\epsilon=
\exp\!\left(
\frac{1}{K}\sum_{k=1}^{K}
\log(|\tilde{\Delta}_k|+\epsilon)
\right),
\]
and assign GeoGain \(+M_\epsilon\) only when all
\(\tilde{\Delta}_k>0\), and \(-M_\epsilon\) otherwise. GeoGain therefore
rewards simultaneous movement in all requested directions while penalizing
imbalanced control. Method comparisons use prompt-paired differences before
averaging within each model--layer setting.
\paragraph{(c) Within-Response Controller Adaptation.}
Prior work on adaptive activation steering has shown the utility of varying
steering intensity during inference
\citep{scalena2024multi,wang2025adaptive,zhao2025adasteer}.
To quantify the realized amount of token-level adaptation in our controller,
we use the range of each target-value coefficient trajectory,
\[
R_{\alpha,k}
=
\max_t \alpha_{k,t}
-
\min_t \alpha_{k,t},
\]
where $\alpha_{k,t}$ denotes the steering coefficient applied to value
$V_k$ at decoding step $t$. For each model, steering objective, and
intervention depth, we average $R_{\alpha,k}$ across target values and
responses. Larger values indicate greater within-response variation, whereas
values near zero indicate that the corresponding coefficient remains nearly
constant throughout generation. This diagnostic characterizes coefficient
variation only and does not, by itself, establish whether that variation is
behaviorally beneficial.
\paragraph{(d) Response Quality.} Stronger behavioral control can potentially come at the cost of generation quality, we independently evaluate whether each steering method
preserves the quality of the generated response. We use a blind GPT-5.6-Sol \citep{openai2026gpt56} judge that observes only the original prompt and generated
response; the steering method, intervention layer, steering objective, and
source foundation are hidden from the judge. The judge independently scores
\emph{coherence} (\(C\)), \emph{fluency} (\(F\)), and \emph{relevance} (\(R\))
on a four-point scale, and we summarize overall response quality as
\[
Q=\frac{C+F+R}{3}.
\]
We report both the raw quality scores and their changes relative to the
corresponding unsteered response,
\(\Delta Q = Q_{\mathrm{steered}}-Q_{\mathrm{unsteered}}\), with analogous
definitions for coherence, fluency, and relevance. Identical prompt--response
pairs are judged only once and their scores are reused across duplicated
conditions, such as the layer-independent Prompt Steering baseline. All
quality comparisons are computed at the exact intervention layer, without
averaging across layers.

\subsection{Robustness Analysis}\label{app:incremental_robustness}
\paragraph{GPT-5.6-Sol: Full depth-wise prompt-level consistency.}
\begingroup
\footnotesize
\setlength{\tabcolsep}{2.8pt}
\renewcommand{\arraystretch}{1.08}
\begin{longtable}{llc cc cc}
 \caption{\textbf{Prompt-level consistency across all intervention depths.} Tie-adjusted \texttt{AIMES} consistency (\%) with pointwise 95\% prompt-bootstrap confidence intervals in brackets ($n=40$ matched prompts per cell).}\\
\toprule
& & & \multicolumn{2}{c}{\textbf{vs. Prompt Steering}} & \multicolumn{2}{c}{\textbf{vs. Fixed Multi}} \\
\cmidrule(lr){4-5}\cmidrule(lr){6-7}
\textbf{Model} & \textbf{Depth} & $\ell$ & $D_x$ & $D_y$  & $D_x$  & $D_y$  \\
\midrule
\endfirsthead
\multicolumn{7}{c}{\tablename\ \thetable{} -- continued} \\
\toprule
& & & \multicolumn{2}{c}{\textbf{vs. Prompt Steering}} & \multicolumn{2}{c}{\textbf{vs. Fixed Multi}} \\
\cmidrule(lr){4-5}\cmidrule(lr){6-7}
\textbf{Model} & \textbf{Depth} & \textbf{L} & $D_x$ Cons. & $D_y$ Cons. & $D_x$ Cons. & $D_y$ Cons. \\
\midrule
\endhead
\midrule
\multicolumn{7}{r}{\textit{Continued on next page}} \\
\endfoot
\bottomrule
\endlastfoot
\multirow{10}{*}{Llama-3.1-8B} & $ $10\% & 3 & 42.5 [28.7, 56.2] & 50.0 [43.8, 57.5] & 41.2 [27.5, 55.0] & 50.0 [43.8, 56.2] \\
 & $ $20\% & 6 & 51.2 [37.5, 65.0] & 50.0 [41.2, 58.8] & 38.7 [25.0, 53.8] & 50.0 [41.2, 58.8] \\
 & $ $30\% & 10 & 55.0 [40.0, 70.0] & 51.2 [43.8, 58.8] & 60.0 [46.2, 73.8] & 52.5 [47.5, 57.5] \\
 & $ $40\% & 13 & 50.0 [36.3, 63.7] & 47.5 [40.0, 53.8] & 36.2 [22.5, 51.2] & 48.8 [42.5, 55.0] \\
 & $ $50\% & 16 & 51.2 [36.2, 66.2] & 55.0 [48.8, 61.3] & 51.2 [38.7, 65.0] & 56.2 [48.8, 63.7] \\
 & $ $60\% & 19 & 53.8 [38.8, 67.5] & 52.5 [43.8, 61.3] & 36.2 [23.8, 50.0] & 52.5 [45.0, 60.0] \\
 & $ $70\% & 22 & 45.0 [31.2, 58.8] & 47.5 [41.2, 53.8] & 50.0 [37.5, 62.5] & 43.8 [36.3, 50.0] \\
 & $ $80\% & 26 & 50.0 [36.2, 63.8] & 50.0 [41.2, 58.8] & 57.5 [43.8, 71.2] & 51.2 [43.8, 58.8] \\
 & $ $90\% & 29 & 60.0 [47.5, 72.5] & 50.0 [42.5, 57.5] & 58.8 [46.2, 71.2] & 47.5 [40.0, 55.0] \\
 & $ $100\% & 32 & 48.8 [36.2, 62.5] & 51.2 [42.5, 58.8] & {65.0 [53.8, 76.2]} & 52.5 [46.2, 58.7] \\
\midrule
\multirow{10}{*}{Gemma-3-4B} & $ $10\% & 3 & 51.2 [37.5, 65.0] & 48.8 [41.2, 56.2] & 40.0 [27.5, 52.5] & 46.2 [38.8, 53.8] \\
 & $ $20\% & 7 & 56.2 [43.8, 68.8] & 53.8 [46.2, 61.3] & 43.8 [30.0, 57.5] & 55.0 [48.8, 62.5] \\
 & $ $30\% & 10 & {67.5 [55.0, 78.8]} & 52.5 [43.8, 61.3] & 62.5 [50.0, 75.0] & 50.0 [43.8, 56.2] \\
 & $ $40\% & 14 & 61.3 [48.8, 73.8] & 48.8 [42.5, 55.0] & 51.2 [38.8, 63.7] & 48.8 [42.5, 55.0] \\
 & $ $50\% & 17 & {67.5 [55.0, 78.8]} & 51.2 [43.8, 58.8] & 53.8 [41.2, 66.2] & 53.8 [48.8, 60.0] \\
 & $ $60\% & 20 & 58.8 [46.2, 71.2] & 53.8 [46.2, 61.3] & 45.0 [32.5, 57.5] & 50.0 [43.8, 57.5] \\
 & $ $70\% & 24 & 57.5 [45.0, 70.0] & 50.0 [42.5, 57.5] & 50.0 [38.7, 61.3] & 47.5 [42.5, 52.5] \\
 & $ $80\% & 27 & 53.8 [41.2, 67.5] & 51.2 [45.0, 57.5] & 45.0 [33.8, 56.2] & 53.8 [50.0, 58.7] \\
 & $ $90\% & 31 & 56.2 [43.8, 68.8] & 51.2 [43.8, 58.8] & 48.8 [38.8, 58.7] & 48.8 [45.0, 52.5] \\
 & $ $100\% & 34 & 58.8 [46.2, 71.2] & 51.2 [45.0, 57.5] & 50.0 [50.0, 50.0] & 50.0 [50.0, 50.0] \\
\midrule
\multirow{10}{*}{Gemma-3-12B} & $ $10\% & 5 & 53.8 [40.0, 67.5] & 55.0 [47.5, 63.7] & 41.2 [31.2, 51.2] & 51.2 [46.2, 56.2] \\
 & $ $20\% & 10 & 55.0 [41.2, 68.8] & 55.0 [46.2, 63.7] & 47.5 [36.2, 58.8] & 48.8 [41.2, 56.2] \\
 & $ $30\% & 14 & 60.0 [46.2, 73.8] & 51.2 [43.8, 60.0] & 48.8 [37.5, 58.8] & 46.2 [40.0, 52.5] \\
 & $ $40\% & 19 & 52.5 [38.8, 66.2] & 50.0 [41.2, 58.8] & 55.0 [43.8, 66.2] & 48.8 [43.8, 53.8] \\
 & $ $50\% & 24 & 56.2 [41.2, 70.0] & 52.5 [45.0, 60.0] & 48.8 [37.5, 60.0] & 55.0 [48.8, 62.5] \\
 & $ $60\% & 29 & 53.8 [40.0, 67.5] & 57.5 [50.0, 65.0] & 48.8 [37.5, 60.0] & 51.2 [46.2, 56.2] \\
 & $ $70\% & 34 & 56.2 [42.5, 70.0] & 55.0 [47.5, 62.5] & 51.2 [40.0, 61.3] & 51.2 [45.0, 57.5] \\
 & $ $80\% & 38 & 57.5 [42.5, 72.5] & 52.5 [45.0, 60.0] & 55.0 [46.2, 63.7] & 48.8 [45.0, 52.5] \\
 & $ $90\% & 43 & 58.8 [45.0, 71.2] & 53.8 [46.2, 61.3] & 53.8 [45.0, 62.5] & 52.5 [47.5, 57.5] \\
 & $ $100\% & 48 & 57.5 [43.8, 71.2] & 52.5 [45.0, 60.0] & 50.0 [50.0, 50.0] & 50.0 [50.0, 50.0] \\
\midrule
\multirow{10}{*}{Qwen3-4B} & $ $10\% & 4 & 53.8 [40.0, 67.5] & 55.0 [47.5, 62.5] & 47.5 [37.5, 57.5] & 51.3 [47.5, 55.0] \\
 & $ $20\% & 7 & 61.3 [47.5, 73.8] & 56.2 [50.0, 62.5] & 46.2 [35.0, 57.5] & 51.2 [45.0, 57.5] \\
 & $ $30\% & 11 & 62.5 [48.8, 76.2] & 52.5 [45.0, 60.0] & 51.2 [40.0, 62.5] & 46.2 [41.2, 51.3] \\
 & $ $40\% & 14 & 61.3 [48.8, 73.8] & 51.2 [43.8, 60.0] & {62.5 [53.8, 71.2]} & 45.0 [38.8, 50.0] \\
 & $ $50\% & 18 & 57.5 [43.8, 71.2] & 52.5 [45.0, 60.0] & 46.2 [35.0, 57.5] & 51.2 [45.0, 57.5] \\
 & $ $60\% & 22 & {65.0 [52.5, 77.5]} & 56.2 [48.8, 63.7] & 60.0 [50.0, 70.0] & 55.0 [50.0, 61.3] \\
 & $ $70\% & 25 & 61.3 [48.8, 73.8] & 53.8 [46.2, 61.3] & 56.2 [48.8, 65.0] & 48.8 [45.0, 52.5] \\
 & $ $80\% & 29 & 61.3 [47.5, 73.8] & 53.8 [46.2, 61.3] & 52.5 [43.8, 61.3] & 50.0 [46.2, 53.8] \\
 & $ $90\% & 32 & 57.5 [43.8, 71.2] & 52.5 [45.0, 58.8] & 55.0 [50.0, 61.3] & 48.8 [46.2, 50.0] \\
 & $ $100\% & 36 & 60.0 [47.5, 72.5] & 53.8 [46.2, 61.3] & 47.5 [42.5, 52.5] & 50.0 [50.0, 50.0] \\
\midrule
\multirow{10}{*}{Qwen3-14B} & $ $10\% & 4 & 55.0 [41.2, 67.5] & 48.8 [40.0, 57.5] & 47.5 [36.3, 58.7] & 52.5 [50.0, 56.2] \\
 & $ $20\% & 8 & 63.8 [50.0, 76.2] & 48.8 [40.0, 57.5] & 53.8 [41.2, 66.3] & 50.0 [45.0, 55.0] \\
 & $ $30\% & 12 & 51.2 [37.5, 65.0] & 50.0 [41.2, 58.8] & 48.8 [38.7, 58.7] & 51.2 [46.2, 56.2] \\
 & $ $40\% & 16 & 55.0 [41.2, 68.8] & 50.0 [41.2, 58.8] & 55.0 [43.8, 66.2] & 47.5 [42.5, 52.5] \\
 & $ $50\% & 20 & 50.0 [36.3, 63.7] & 48.8 [40.0, 57.5] & 45.0 [33.8, 57.5] & 48.8 [43.8, 53.8] \\
 & $ $60\% & 24 & 56.2 [43.8, 67.5] & 51.2 [43.8, 60.0] & 51.2 [40.0, 62.5] & 51.2 [46.2, 56.2] \\
 & $ $70\% & 28 & 56.2 [42.5, 70.0] & 52.5 [43.8, 61.3] & 52.5 [42.5, 62.5] & 53.8 [50.0, 58.7] \\
 & $ $80\% & 32 & 58.8 [45.0, 71.2] & 48.8 [40.0, 57.5] & 48.8 [40.0, 58.8] & 51.3 [47.5, 55.0] \\
 & $ $90\% & 36 & 55.0 [42.5, 68.8] & 50.0 [41.2, 58.8] & 40.0 [32.5, 47.5] & 51.2 [50.0, 53.8] \\
 & $ $100\% & 40 & 55.0 [42.5, 67.5] & 50.0 [41.2, 58.8] & 47.5 [43.8, 50.0] & 48.8 [46.2, 50.0]
\label{tab:aimes_consistency_full} 
\end{longtable}

\endgroup

\paragraph{Claude-Opus-4.8: Full depth-wise prompt-level consistency.}
Table~\ref{tab:claude_consistency_full} complements the Claude mean interaction contrasts in Figure~\ref{fig:claude_depth_consistency} by showing how consistently the corresponding advantage occurs across individual prompts at every tested intervention depth. Each entry reports the tie-adjusted percentage of matched prompts for which \texttt{AIMES} obtains a larger $D_x$ or $D_y$ interaction effect than the corresponding comparator. Values above $50\%$ therefore indicate that the \texttt{AIMES} advantage occurs for a majority of prompts. The results show that prompt-level consistency varies across models, objectives, and intervention depths, while positive-majority consistency appears across a broad range of settings for both comparison methods.

\begingroup
\footnotesize
\setlength{\tabcolsep}{2.8pt}
\renewcommand{\arraystretch}{1.08}
\begin{longtable}{llc cc cc}
\caption{\textbf{Claude-Opus-4.8 prompt-level consistency across all intervention depths.} Tie-adjusted \texttt{AIMES} consistency (\%) under Claude-Opus-4.8 with pointwise 95\% prompt-bootstrap confidence intervals in brackets ($n=40$ matched prompts per cell). $D_x$ measures Care/Fairness retention and $D_y$ additional Sanctity suppression.}\\
\toprule
& & & \multicolumn{2}{c}{\textbf{vs. Prompt Steering}} & \multicolumn{2}{c}{\textbf{vs. Fixed Multi}} \\
\cmidrule(lr){4-5}\cmidrule(lr){6-7}
\textbf{Model} & \textbf{Depth} & $\ell$ & $D_x$ & $D_y$ & $D_x$ & $D_y$ \\
\midrule
\endfirsthead
\multicolumn{7}{c}{\tablename\ \thetable{} -- continued} \\
\toprule
& & & \multicolumn{2}{c}{\textbf{vs. Prompt Steering}} & \multicolumn{2}{c}{\textbf{vs. Fixed Multi}} \\
\cmidrule(lr){4-5}\cmidrule(lr){6-7}
\textbf{Model} & \textbf{Depth} & $\ell$ & $D_x$ & $D_y$ & $D_x$ & $D_y$ \\
\midrule
\endhead
\midrule
\multicolumn{7}{r}{\textit{Continued on next page}} \\
\endfoot
\bottomrule
\endlastfoot
\multirow{10}{*}{Llama-3.1-8B} & $10\%$ & 3 & 31.2 [21.2, 42.5] & 48.8 [41.2, 56.2] & 45.0 [33.8, 56.2] & 50.0 [43.8, 57.5] \\
 & $20\%$ & 6 & 43.8 [31.2, 56.2] & 48.8 [41.2, 56.2] & 45.0 [33.8, 56.2] & 47.5 [41.2, 53.8] \\
 & $30\%$ & 10 & 48.8 [36.3, 61.3] & {55.0} [47.5, 62.5] & {55.0} [43.8, 66.2] & {51.2} [45.0, 57.5] \\
 & $40\%$ & 13 & 42.5 [31.2, 53.8] & {55.0} [48.8, 62.5] & {52.5} [38.8, 66.2] & {56.2} [51.2, 61.3] \\
 & $50\%$ & 16 & 42.5 [31.2, 53.8] & {51.2} [43.8, 58.8] & {53.8} [42.5, 65.0] & {52.5} [46.2, 58.8] \\
 & $60\%$ & 19 & 45.0 [32.5, 57.5] & 47.5 [40.0, 55.0] & 48.8 [37.5, 58.8] & 45.0 [37.5, 51.2] \\
 & $70\%$ & 22 & 41.2 [28.7, 53.8] & {52.5} [46.2, 60.0] & {53.8} [42.5, 65.0] & 50.0 [43.8, 56.2] \\
 & $80\%$ & 26 & 43.8 [32.5, 55.0] & 50.0 [42.5, 57.5] & 41.2 [31.2, 51.2] & 47.5 [41.2, 53.8] \\
 & $90\%$ & 29 & {51.2} [38.8, 63.7] & {53.8} [46.2, 61.3] & {56.2} [46.2, 65.0] & {51.2} [46.2, 56.2] \\
 & $100\%$ & 32 & 50.0 [37.5, 62.5] & {51.2} [45.0, 57.5] & {55.0} [46.2, 63.7] & 50.0 [45.0, 55.0] \\
\midrule
\multirow{10}{*}{Gemma-3-4B} & $10\%$ & 3 & {58.8} [48.8, 68.8] & 48.8 [42.5, 53.8] & {52.5} [42.5, 62.5] & {51.2} [46.2, 56.2] \\
 & $20\%$ & 7 & {53.8} [43.8, 63.7] & 48.8 [42.5, 53.8] & {56.2} [48.8, 63.7] & 48.8 [43.8, 53.8] \\
 & $30\%$ & 10 & {53.8} [43.8, 62.5] & 48.8 [42.5, 55.0] & {55.0} [47.5, 62.5] & 50.0 [45.0, 55.0] \\
 & $40\%$ & 14 & {52.5} [42.5, 62.5] & 48.8 [43.8, 53.8] & 47.5 [38.7, 56.2] & 48.8 [43.8, 53.8] \\
 & $50\%$ & 17 & {52.5} [43.8, 61.3] & 47.5 [41.2, 53.8] & 48.8 [40.0, 56.2] & {51.2} [46.2, 56.2] \\
 & $60\%$ & 20 & 50.0 [41.2, 58.8] & {51.2} [45.0, 57.5] & 46.2 [38.7, 53.8] & {52.5} [46.2, 58.8] \\
 & $70\%$ & 24 & {51.2} [40.0, 62.5] & 48.8 [43.8, 53.8] & {51.2} [41.2, 61.3] & 50.0 [46.2, 53.8] \\
 & $80\%$ & 27 & {53.8} [45.0, 62.5] & {51.2} [46.2, 56.2] & {56.2} [48.8, 63.7] & 46.2 [41.2, 50.0] \\
 & $90\%$ & 31 & {53.8} [45.0, 62.5] & 48.8 [42.5, 53.8] & {53.8} [47.5, 60.0] & 48.8 [46.2, 50.0] \\
 & $100\%$ & 34 & {53.8} [45.0, 62.5] & 48.8 [43.8, 53.8] & 47.5 [41.2, 53.8] & 50.0 [50.0, 50.0] \\
\midrule
\multirow{10}{*}{Gemma-3-12B} & $10\%$ & 5 & {57.5} [47.5, 67.5] & {55.0} [48.8, 62.5] & 50.0 [42.5, 57.5] & 48.8 [45.0, 52.5] \\
 & $20\%$ & 10 & {52.5} [42.5, 62.5] & {55.0} [48.8, 62.5] & 48.8 [40.0, 57.5] & {51.3} [47.5, 55.0] \\
 & $30\%$ & 14 & {55.0} [45.0, 65.0] & {55.0} [50.0, 61.3] & 50.0 [42.5, 57.5] & 47.5 [43.8, 50.0] \\
 & $40\%$ & 19 & {53.8} [43.8, 63.7] & {55.0} [50.0, 61.3] & 48.8 [40.0, 56.2] & 50.0 [45.0, 55.0] \\
 & $50\%$ & 24 & {55.0} [45.0, 65.0] & {53.8} [47.5, 60.0] & {51.2} [46.2, 56.2] & 50.0 [46.2, 53.8] \\
 & $60\%$ & 29 & {57.5} [46.2, 67.5] & {58.8} [52.5, 66.2] & {57.5} [50.0, 65.0] & {53.8} [50.0, 58.7] \\
 & $70\%$ & 34 & {53.8} [43.8, 63.7] & {52.5} [46.2, 60.0] & {51.2} [43.8, 58.8] & 46.2 [41.2, 50.0] \\
 & $80\%$ & 38 & {55.0} [45.0, 65.0] & {56.2} [50.0, 62.5] & 50.0 [43.8, 56.2] & {52.5} [50.0, 56.2] \\
 & $90\%$ & 43 & {55.0} [45.0, 65.0] & {55.0} [48.8, 62.5] & 48.8 [43.8, 52.5] & 48.8 [45.0, 52.5] \\
 & $100\%$ & 48 & {52.5} [43.8, 61.3] & {57.5} [51.2, 63.7] & 45.0 [40.0, 48.8] & {52.5} [50.0, 56.2] \\
\midrule
\multirow{10}{*}{Qwen3-4B} & $10\%$ & 4 & 41.2 [32.5, 50.0] & {51.2} [45.0, 57.5] & 42.5 [33.8, 50.0] & {51.2} [47.5, 55.0] \\
 & $20\%$ & 7 & {52.5} [42.5, 62.5] & {52.5} [46.2, 58.8] & {53.8} [45.0, 62.5] & 50.0 [46.2, 53.8] \\
 & $30\%$ & 11 & {58.8} [48.8, 68.8] & {51.2} [46.2, 56.2] & {63.7} [56.2, 71.2] & 46.2 [41.2, 50.0] \\
 & $40\%$ & 14 & 46.2 [36.2, 56.2] & {51.2} [45.0, 57.5] & {55.0} [46.2, 63.7] & 46.2 [41.2, 50.0] \\
 & $50\%$ & 18 & 43.8 [35.0, 52.5] & {55.0} [48.8, 62.5] & 45.0 [36.3, 53.8] & {52.5} [47.5, 57.5] \\
 & $60\%$ & 22 & {51.2} [41.2, 61.3] & 50.0 [43.8, 56.2] & {55.0} [46.2, 63.7] & 50.0 [43.8, 56.2] \\
 & $70\%$ & 25 & 47.5 [37.5, 57.5] & {53.8} [47.5, 60.0] & 50.0 [42.5, 57.5] & {52.5} [50.0, 56.2] \\
 & $80\%$ & 29 & 48.8 [40.0, 58.7] & {52.5} [46.2, 58.8] & {51.2} [43.8, 58.8] & {52.5} [50.0, 56.2] \\
 & $90\%$ & 32 & 47.5 [38.8, 56.2] & {52.5} [46.2, 58.7] & {52.5} [50.0, 56.2] & 48.8 [46.2, 50.0] \\
 & $100\%$ & 36 & 50.0 [41.2, 58.8] & {52.5} [46.2, 58.7] & {51.2} [47.5, 55.0] & 50.0 [50.0, 50.0] \\\\
\midrule
\multirow{10}{*}{Qwen3-14B} & $10\%$ & 4 & {57.5} [45.0, 70.0] & {52.5} [45.0, 60.0] & 48.8 [41.2, 56.2] & {51.2} [47.5, 55.0] \\
 & $20\%$ & 8 & {55.0} [43.8, 66.3] & {53.8} [47.5, 60.0] & {55.0} [46.2, 63.7] & {51.2} [47.5, 55.0] \\
 & $30\%$ & 12 & {56.2} [43.8, 68.8] & {55.0} [48.8, 62.5] & {55.0} [47.5, 62.5] & {51.3} [47.5, 56.2] \\
 & $40\%$ & 16 & {53.8} [42.5, 65.0] & {52.5} [46.2, 58.7] & 45.0 [37.5, 51.2] & 50.0 [46.2, 53.8] \\
 & $50\%$ & 20 & {55.0} [43.8, 66.2] & {52.5} [46.2, 58.7] & 50.0 [43.8, 56.2] & 46.2 [41.2, 50.0] \\
 & $60\%$ & 24 & {53.8} [42.5, 65.0] & {53.8} [47.5, 60.0] & 48.8 [43.8, 53.8] & 50.0 [46.2, 53.8] \\
 & $70\%$ & 28 & {58.8} [46.2, 70.0] & {52.5} [46.2, 58.8] & {52.5} [46.2, 58.8] & 50.0 [46.2, 53.8] \\
 & $80\%$ & 32 & {55.0} [45.0, 65.0] & {52.5} [46.2, 58.8] & 50.0 [43.8, 56.2] & {51.2} [50.0, 53.8] \\
 & $90\%$ & 36 & {57.5} [46.2, 68.8] & {52.5} [46.2, 58.8] & {55.0} [50.0, 61.3] & 48.8 [46.2, 50.0] \\
 & $100\%$ & 40 & {55.0} [43.8, 66.2] & {52.5} [46.2, 58.7] & {51.2} [45.0, 57.5] & 50.0 [50.0, 50.0] \\
 \label{tab:claude_consistency_full} 
\end{longtable}
\endgroup
\normalsize
\subsection{Detailed Response-Quality Results}
\label{app:response_quality}

\begin{table}[h]
\small
\centering
\resizebox{\columnwidth}{!}{
\begin{tabular}{lccc}
\toprule
\textbf{Objective}
&
\textbf{\texttt{AIMES} $>$ Fixed}
&
\textbf{\texttt{AIMES} $>$ Prompt}
&
\textbf{\texttt{AIMES} $>$ Prompt} \\
&
\textbf{on $Q$}
&
\textbf{on $Q$}
&
\textbf{on Relevance} \\
\midrule

$+\,$Care $+\,$Fairness
&
26/50
&
29/50
&
41/50
\\

$+\,$Loyalty $+\,$Authority
&
24/50
&
22/50
&
30/50
\\

$+\,$Care $+\,$Fairness $-\,$Sanctity
&
26/50
&
30/50
&
35/50
\\

\midrule

\textbf{Overall}
&
\textbf{76/150}
&
\textbf{81/150}
&
\textbf{106/150}
\\

\bottomrule
\end{tabular}
}
\caption{\textbf{Blind response-quality comparison.}
Number of model--depth settings in which \texttt{AIMES} attains a higher
mean score than each comparator. Overall quality is defined as
\(Q=(C+F+R)/3\), where \(C\), \(F\), and \(R\) denote coherence, fluency,
and relevance. Counts are descriptive rather than significance tests.}
\label{tab:response_quality}
\end{table}

We evaluate whether the multi-value control gains of \texttt{AIMES} are
associated with changes in generation quality using a blind
\texttt{GPT-5.6-Sol} judge. Each response is scored for coherence (\(C\)),
fluency (\(F\)), and relevance (\(R\)) on a four-point scale, with overall
quality defined as \(Q=(C+F+R)/3\).
Table~\ref{tab:response_quality} summarizes the number of
model--depth--objective settings in which \texttt{AIMES} attains a higher
mean score than each comparator. Across the 150 settings, \texttt{AIMES}
exceeds Fixed Multi-Value Steering on \(Q\) in \(76\) settings and Prompt Steering in
\(81\), indicating broadly comparable overall response quality rather than a
consistent quality advantage. For relevance, \texttt{AIMES} exceeds Prompt
Steering in \(106/150\) settings (\(70.7\%\)). Taken together, these results
do not indicate a systematic loss in response quality under adaptive steering.

\subsection{Layer-wise Within-Response Steering Variation}
\label{app:controller_adaptation}
\begin{figure}[h]
    \centering
    \includegraphics[scale=0.35]{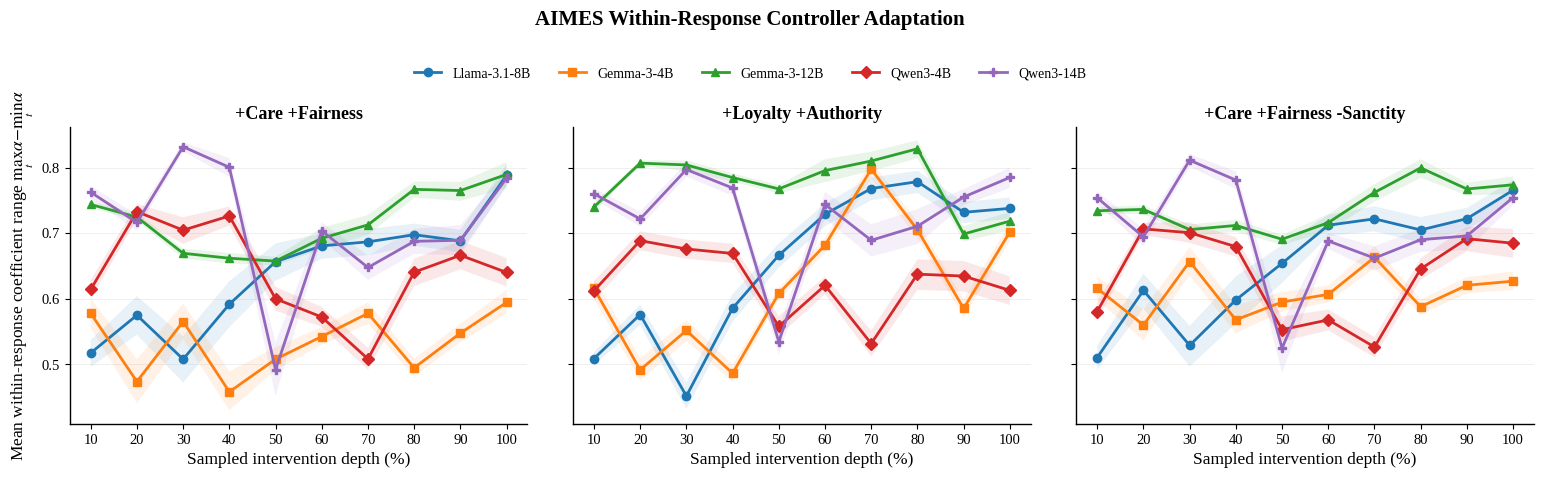}
    \caption{\textbf{Variation in \texttt{AIMES} steering strength across intervention depths.}
    For each model and objective, we measure the within-response range of the target-value steering coefficients,
    $R_{\alpha,k}=\max_t\alpha_{k,t}-\min_t\alpha_{k,t}$.
    Each point shows the mean coefficient range across target-disjoint responses at the corresponding model-specific intervention layer, and shaded regions denote $95\%$ prompt-bootstrap confidence intervals. Larger values indicate greater changes in steering strength during generation. Results are reported separately at each of the ten sampled depths; no averaging is performed across intervention layers.}
    \label{fig:app:controller}
\end{figure}
A defining feature of \texttt{AIMES} is that its steering coefficients are updated during generation using the online value readout, rather than remaining fixed for the entire response. We therefore examine how much these coefficients vary within a response across the full intervention-depth grid.

\paragraph{Metric.}
For each controlled value $V_k$, we measure the within-response coefficient range
\begin{equation}
R_{\alpha,k}
=
\max_t \alpha_{k,t}
-
\min_t \alpha_{k,t},
\end{equation}
where $\alpha_{k,t}$ is the steering coefficient applied to value $V_k$ at decoding step $t$. For objectives involving multiple target values, we first average $R_{\alpha,k}$ across the controlled values within each response and then average across target-disjoint prompts. Thus, larger values indicate greater changes in steering strength over the course of generation, while smaller values indicate more nearly constant steering. The analysis is performed independently at each sampled intervention layer; no averaging is performed across layers.

\paragraph{Depth-wise analysis.}
Figure~\ref{fig:app:controller} reports the resulting coefficient variation across all ten sampled intervention depths ($10\%$--$100\%$) for the three steering objectives. Across all five models, the coefficient range remains clearly nonzero throughout the network, showing that \texttt{AIMES} continues to adjust its steering strength during generation rather than reducing to a fixed-coefficient intervention. The magnitude of this variation depends on the model, objective, and intervention depth, as expected from a controller whose updates are driven by the model-specific online value state.

The same qualitative behavior is observed across both two-value objectives,
$(\uparrow\mathrm{Care}\;\uparrow\mathrm{Fairness})$ and
$(\uparrow\mathrm{Loyalty}\;\uparrow\mathrm{Authority})$, as well as the competing three-value objective
$(\uparrow\mathrm{Care}\;\uparrow\mathrm{Fairness}\;\downarrow\mathrm{Sanctity})$.
Across the plotted conditions, mean coefficient ranges are typically substantial, with values approximately spanning $0.45$--$0.84$. Importantly, this variation is not confined to a single model family or a narrow depth region: all five models exhibit clear within-response changes across early, middle, and later intervention layers.

The representative $30\%$ depth reported in the main text is therefore not an isolated case. Rather, it is part of a broader depth-wise pattern in which the adaptive coefficients remain responsive throughout generation. At this depth, the coefficient range is consistently nonzero across all models and objectives, while the full analysis here shows that this behavior persists across the complete set of sampled intervention layers.

Overall, the layer-wise results confirm that the adaptive component of \texttt{AIMES} remains active across the network: its steering coefficients change meaningfully within responses across models, objectives, and intervention depths, rather than behaving as approximately fixed steering weights.

\subsection{Independent-Judge-based Analysis of Depth-wise and GeoGain Results}
\label{app:external_judge}
\begin{figure}[t]
    \centering
    \includegraphics[scale=0.32]{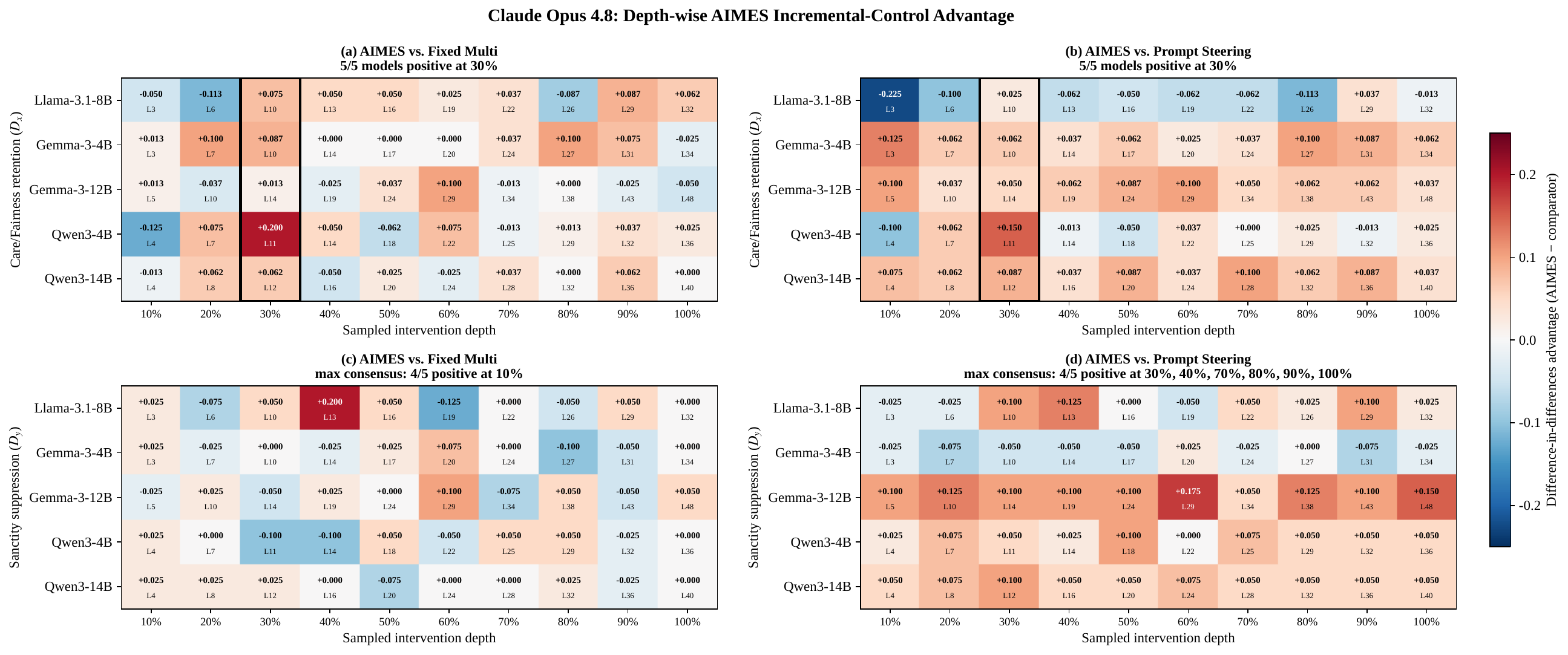}
    \caption{\textbf{Claude-Opus-4.8 judge-based analysis of the depth-wise \texttt{AIMES} advantage.}
    The analysis uses the same ten sampled intervention depths and the same $D_x/D_y$ definitions as Figure~\ref{fig:depth_consistency}, but re-scores the identical generations with Claude-Opus-4.8. Positive values indicate an advantage for \texttt{AIMES}; black outlines denote depths at which the advantage is positive for all five models.}
    \label{fig:claude_depth_consistency}
\end{figure}

\paragraph{Depth-wise interaction contrasts.}
Figure~\ref{fig:claude_depth_consistency} reports the Claude-based depth-wise advantage of \texttt{AIMES} using the same interaction contrasts as Figure~\ref{fig:depth_consistency}. The Care/Fairness-retention advantage over Fixed Multi-Value Steering, $D_x$, reproduces the strongest depth-specific pattern observed under GPT-5.6-Sol: at $30\%$ depth, $D_x>0$ for all five models under both judges. The corresponding comparison against Prompt Steering also exhibits 5/5 positive agreement at $30\%$ depth, although the additional depths identified as unanimous under GPT-5.6-Sol do not all remain unanimous under Claude. For example, at $80\%$ depth, Llama-3.1-8B changes sign under Claude ($D_x=-0.113$).
The Sanctity-suppression contrast, $D_y$, shows greater evaluator sensitivity than the
Care/Fairness-retention contrast. Against Fixed Multi-Value Steering, the depth at which the advantage concentrates differs between evaluators: under GPT-5.6-Sol, the strongest
consensus occurs near $60\%$ depth, whereas under Claude, the strongest consensus for this comparator occurs near $10\%$ depth (Figure~\ref{fig:claude_depth_consistency} (c)). By contrast, the comparison with
Prompt Steering is more consistent across evaluators, remaining broadly positive with four of five models positive at several depths under both judges, though without a fully unanimous column under either. Overall, the full-resolution analysis indicates that the $30\%$
Care/Fairness-retention advantage over Fixed Multi-Value Steering is the most evaluator-robust depth-specific finding in this analysis; the Sanctity-suppression advantage over Fixed Multi-Value Steering is depth-dependent in a way that does not consistently localize to the same network
region across evaluators, and we scope our main-text claim about this comparison accordingly (Section~\ref{sec:experiments}).
\begin{figure}[t]
    \centering
    \includegraphics[width=\textwidth]{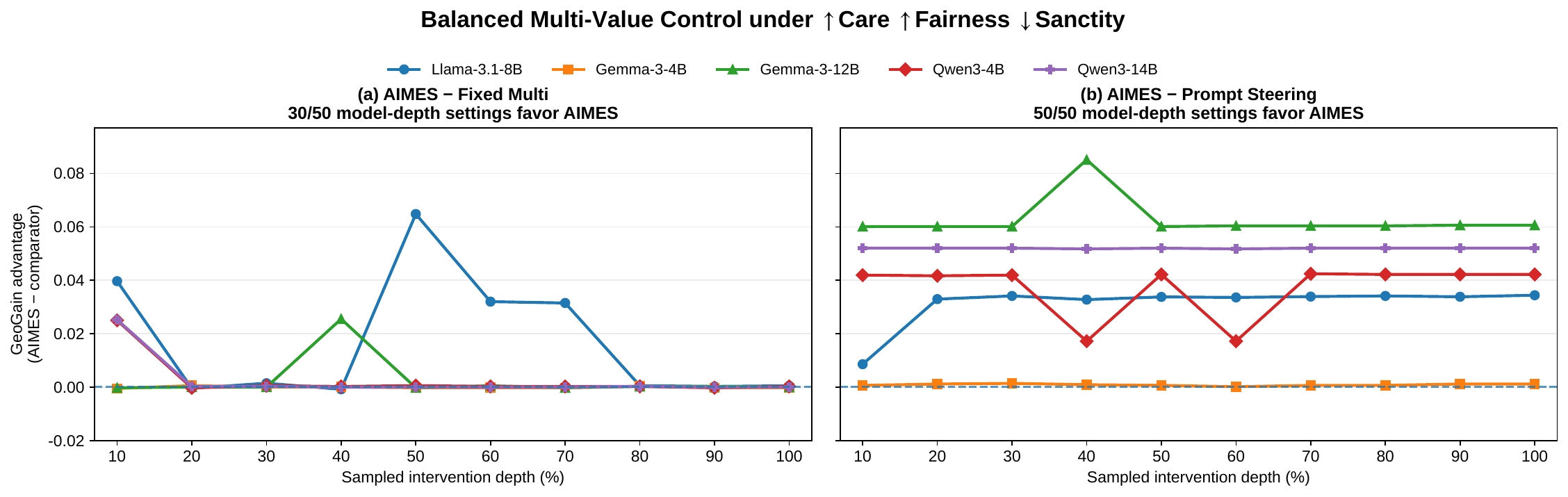}
    \caption{\textbf{Claude-Opus-4.8 judge-based analysis of GeoGain for the competing three-value objective.}
    Prompt-paired GeoGain advantage of \texttt{AIMES} over Fixed Multi-Value Steering (left) and Prompt Steering (right) across all ten sampled intervention depths for
    $(\uparrow\mathrm{Care}\;\uparrow\mathrm{Fairness}\;\downarrow\mathrm{Sanctity})$.
    Positive values indicate higher GeoGain metric for \texttt{AIMES} vs. Fixed Multi-Value Steering and Prompt Steering. Under Claude-Opus-4.8, 30/50 model-depth settings favor \texttt{AIMES} over Fixed Multi-Value Steering and 50/50 favor \texttt{AIMES} over Prompt Steering.}
    \label{fig:claude_geogain}
\end{figure}
\paragraph{GeoGain Analysis.}
We next recompute the prompt-level GeoGain analysis using Claude scores over the complete ten-depth grid. Table~\ref{tab:claude_geogain_replication} summarizes the number of model--depth cells in which \texttt{AIMES} attains higher GeoGain than each comparator. Across the three objectives and two comparators, this yields $3\times2\times50=300$ model--depth comparisons.

Relative to Fixed Multi-Value Steering, \texttt{AIMES} attains higher GeoGain in 35/50 settings for
$(\uparrow\mathrm{Care}\;\uparrow\mathrm{Fairness})$, 29/50 for
$(\uparrow\mathrm{Loyalty}\;\uparrow\mathrm{Authority})$, and 30/50 for
$(\uparrow\mathrm{Care}\;\uparrow\mathrm{Fairness}\;\downarrow\mathrm{Sanctity})$.
The latter exactly matches the 30/50 positive model--depth count obtained with GPT-5.6-Sol, providing independent-judge support for the model- and depth-dependent GeoGain advantage over Fixed Multi-Value Steering. Importantly, the 30/50 count understates the asymmetry in effect magnitude: on the three-value objective, cells in which \texttt{AIMES} trails Fixed Multi-Value Steering differ by at most approximately $-0.001$ in GeoGain, whereas favorable cells reach advantages as large as approximately $+0.065$. Thus, the non-positive cells are predominantly near ties, while several of the positive cells exhibit substantially larger margins.
\begin{table}[h]
\centering
\begin{tabular}{lcc}
\toprule
Objective & \texttt{AIMES} vs.\ Fixed Multi & \texttt{AIMES} vs.\ Prompt Steering \\
\midrule
$(\uparrow\mathrm{Care}\;\uparrow\mathrm{Fairness})$
& 35/50 & 0/50 \\
$(\uparrow\mathrm{Loyalty}\;\uparrow\mathrm{Authority})$
& 29/50 & 21/50 \\
$(\uparrow\mathrm{Care}\;\uparrow\mathrm{Fairness}\;\downarrow\mathrm{Sanctity})$
& 30/50 & \textbf{50/50} \\
\bottomrule
\end{tabular}
\caption{\textbf{Claude-Opus-4.8 GeoGain Analysis.}
Number of model--depth cells, out of 50 (five models $\times$ ten depths), in which \texttt{AIMES} attains higher prompt-level GeoGain than each comparator.}
\label{tab:claude_geogain_replication}
\end{table}
The comparison with Prompt Steering is strongly objective-dependent. For
$(\uparrow\mathrm{Care}\;\uparrow\mathrm{Fairness})$, \texttt{AIMES} has lower GeoGain in all 50 model--depth settings, whereas for
$(\uparrow\mathrm{Loyalty}\;\uparrow\mathrm{Authority})$, it is higher in 21/50 settings. In contrast, for the competing three-value objective
$(\uparrow\mathrm{Care}\;\uparrow\mathrm{Fairness}\;\downarrow\mathrm{Sanctity})$,
\texttt{AIMES} attains higher GeoGain than Prompt Steering in \textbf{50/50} model--depth settings. This strengthens the near-universal advantage observed with GPT-5.6-Sol, where 47/50 settings favored \texttt{AIMES}, and shows that the principal balanced-control result persists under an independent evaluator across the full depth grid. Because GeoGain is positive only when all requested target directions improve simultaneously, the 50/50 \texttt{AIMES} advantage on the three-value objective indicates more favorable joint satisfaction of the competing target directions under Claude, rather than simply larger movement along any individual value dimension.

The model-wise curves further illustrate this reversal. Gemma-3-12B exhibits the largest \texttt{AIMES} advantage over Prompt Steering on the three-value objective, approximately $+0.06$ to $+0.085$ across depth, despite \texttt{AIMES} being disadvantaged for this model on the simpler two-value objectives. This suggests that the relative advantage emerges specifically when an additional competing constraint must be satisfied jointly, rather than from uniformly stronger steering on individual target values.

Taken together, the Claude judge-based analysis supports the main GeoGain conclusion while revealing greater evaluator sensitivity in the finer-grained depth-wise decomposition. In particular, the $30\%$ Care/Fairness-retention effect and the balanced-control advantage on the three-value objective persist across evaluators, whereas the precise depth at which Sanctity-suppression advantages emerge is less stable.

\end{document}